\documentclass[10pt,twocolumn,letterpaper]{article}

\usepackage[pagenumbers]{cvpr} % To force page numbers, e.g. for an arXiv version

\usepackage{graphicx}
\usepackage{amsmath}
\usepackage{amssymb}
\usepackage{booktabs}

\usepackage[pagebackref,breaklinks,colorlinks]{hyperref}

\usepackage[accsupp]{axessibility}

\usepackage{multirow}
\usepackage[capitalize]{cleveref}
\crefname{section}{Sec.}{Secs.}
\Crefname{section}{Section}{Sections}
\Crefname{table}{Table}{Tables}
\crefname{table}{Tab.}{Tabs.}

\def\cvprPaperID{10737} % *** Enter the CVPR Paper ID her

\def\confName{CVPR}
\def\confYear{2023}

\begin{document}

\title{Masked Wavelet Representation for Compact Neural Radiance Fields}

\newcommand\CoAuthorMark{\footnotemark[\arabic{footnote}]}
\newcommand\CorrespondingAuthorMark{\footnotemark[\arabic{footnote}]}
\author{Daniel Rho$^1$\thanks{Equal contribution} \and
Byeonghyeon Lee$^2$\protect\CoAuthorMark \and
Seungtae Nam$^2$ \and
Joo Chan Lee$^2$ \and
Jong Hwan Ko$^{2,3}$\thanks{Corresponding authors}\and
Eunbyung Park$^{2,3}$\protect\CorrespondingAuthorMark \\
$^1$AI2XL, KT \\
$^2$Department of Artificial Intelligence, Sungkyunkwan University \\
$^3$Department of Electrical and Computer Engineering, Sungkyunkwan University}

\maketitle

%%%%%%%%% ABSTRACT
\begin{abstract}
Neural radiance fields (NeRF) have demonstrated the potential of coordinate-based neural representation (neural fields or implicit neural representation) in neural rendering.
However, using a multi-layer perceptron (MLP) to represent a 3D scene or object requires enormous computational resources and time.
There have been recent studies on how to reduce these computational inefficiencies by using additional data structures, such as grids or trees.
Despite the promising performance, the explicit data structure necessitates a substantial amount of memory.
In this work, we present a method to reduce the size without compromising the advantages of having additional data structures.
In detail, we propose using the wavelet transform on grid-based neural fields.
Grid-based neural fields are for fast convergence, and the wavelet transform, whose efficiency has been demonstrated in high-performance standard codecs, is to improve the parameter efficiency of grids.
Furthermore, in order to achieve a higher sparsity of grid coefficients while maintaining reconstruction quality, we present a novel trainable masking approach.
Experimental results demonstrate that non-spatial grid coefficients, such as wavelet coefficients, are capable of attaining a higher level of sparsity than spatial grid coefficients, resulting in a more compact representation.
With our proposed mask and compression pipeline, we achieved state-of-the-art performance within a memory budget of 2 MB.
Our code is available at \href{https://github.com/daniel03c1/masked\_wavelet\_nerf}{https://github.com/daniel03c1/masked\_wavelet\_nerf}.
\end{abstract}

%%%%%%%%% BODY TEXT
\section{Introduction}
\label{sec:intro}

\begin{figure}[t]
\begin{center}
\includegraphics[width=1.0\linewidth]{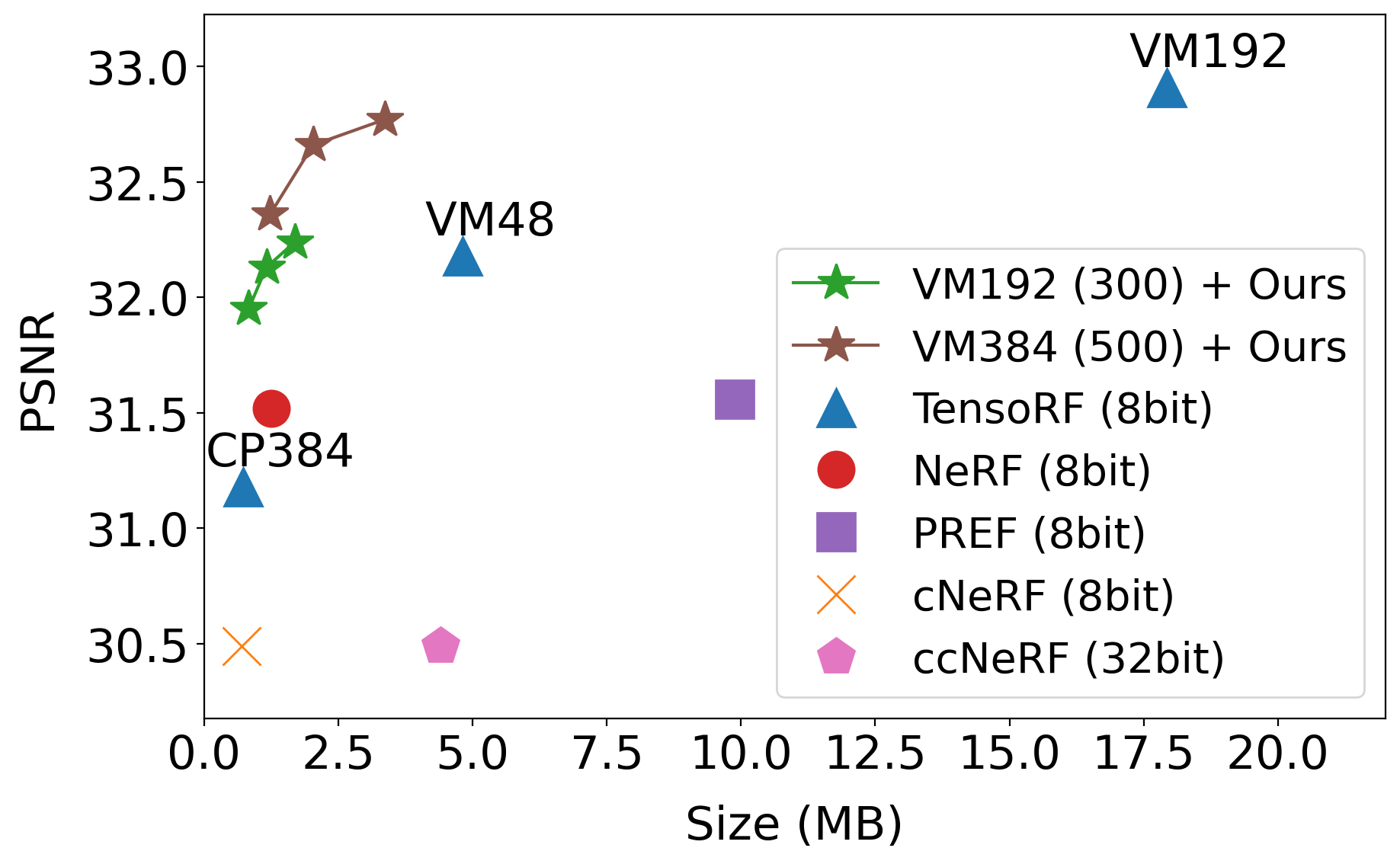}
\end{center}
\vspace{-0.5cm}
   \caption{Rate-distortion curves on the NeRF synthetic dataset. The numbers inside parenthesis denote the axis resolution of grids.}
\label{fig:main-rate-distortion}
\end{figure}

% NeRF as a new media paradigm
Recent advances in coordinate-based neural representation (neural fields or implicit neural representation) have demonstrated remarkable performance in many applications.
In particular, neural radiance fields (NeRF) have sparked interest by synthesizing high-quality images from novel viewpoints.
It uses a multi-layer perceptron (MLP) with positional encoding to map coordinates to corresponding colors and opacities.
Combined with the differentiable volumetric rendering and the neural network's architectural priors, it has shown great potential to be a new representation paradigm.
However, the high computational costs (in both training and inference) have been a significant bottleneck, often taking several days to converge.

% making faster, but more space or meta-learning needs pretraining and poor generalization performance.
Several follow-up studies have been proposed to accelerate training and inference times~\cite{kilonerf,FastNeRF,learned-init,TiNeuVox,R2L,EfficientNeRF,mvsnerf,dsnerf,derf}.
To speed up inference, KiloNeRF~\cite{kilonerf} proposed splitting a 3D scene into thousands of partial scenes, each of which is assigned a tiny, distinct neural network. % KiloNeRF~\cite{kilonerf} proposed splitting a 3D scene into thousands of partial scenes allocated to tiny individual networks to speed up inference.
While achieving impressive speed-up, it requires a massive amount of memory storage.
On the other hand, FastNeRF~\cite{FastNeRF} suggested caching and factorizing the NeRF network to reduce computational costs.
Meanwhile, meta-learning algorithms~\cite{maml,reptile,anil} have also been applied to accelerate the training time, and they have shown faster convergence with the learned initialization~\cite{learned-init,MetaSDF}. % to this task for faster convergence at training time with learned initial weights~\cite{learned-init}, leveraging a large-scale dataset.
% While achieving impressive speed-up, the former two approaches require a massive amount of storage.
However, it requires well-organized large-scale datasets and a pre-training process.
%Neural fields might converge faster with learned initial weights, but at the cost of well-organized large-scale datasets and a pre-training process.
% In addition, it suffers from out-of-distribution generalization.
% This approach necessitates well-organized large-scale datasets and a pre-training process and suffers from out-of-distribution generalization.
% Furthermore, because meta-learning can only reduce the number of training iterations, the rendering costs remain unchanged.
Furthermore, the rendering costs remain unchanged since they use the same network architecture~\cite{learned-init}, and meta-learning algorithms often suffer from poor out-of-distribution generalization performance.
% Most importantly, only the training speed could be sped up while the inference speed remained constant.

% better speed, less storage, high quality with grid
Alternatively, there has been a surge of recent interest in incorporating classical data structures, such as grids or trees, into the NeRF framework~\cite{NSVF, Plenoxels,PlenOctrees,instant-ngp,VQAD, nglod,schwarz2022voxgraf}.
Incorporating additional data structures has significantly reduced training and inference time (from days to a few minutes) without compromising reconstruction quality.
However, the overall size dramatically increases due to these dense and volumetric structures.
%However, the overall size is increased by these dense and volumetric structures. % One drawback of this approach is that the overall size increases.
In order to reduce the spatial complexity, several methods have been proposed, including pruning areas~\cite{PlenOctrees,DVGO,schwarz2022voxgraf}, encodings~\cite{instant-ngp,VQAD}, and tensor decomposition~\cite{EG3D,tensorf,huang2022pref}.
% These methods simultaneously achieved compact representation, quick training and inference times, and high reconstruction quality. % , achieving compact representation, rapid training time, and high reconstruction quality simultaneously.

% DWT
This paper aims to further improve the spatial complexity while maintaining the rendering quality.
Leveraging the decades of research on standard compression algorithms~\cite{jpeg2000,h264,hevc}, we propose compressing grid-based neural fields using frequency-based transformations.
In the frequency domain, a large portion of the coefficients can be discarded without considerably degrading the reconstruction quality, and most standard compression algorithms have exploited frequency domain representations.
%Because of this, most compression algorithms have used frequency domain representations. % This advantage of signal representation in the frequency domain has long been exploited by most compression algorithms.
Thus, we propose using this property on grid-based neural fields to maximize parameter efficiency. % in the grid-based representation.
Among other alternatives, we employ the discrete wavelet transform (DWT) due to its compactness and ability to efficiently capture both global and local information.

% Mask
Once we obtain sparse representations via frequency domain representations, we can take advantage of the existing compression techniques.
Unlike conventional media data (e.g., image and audio), however, no off-the-shelf compression tools exist that we can leverage without complicated engineering efforts.
In addition, since NeRF, or neural rendering networks in general, is a relatively new data format, the characteristics or patterns of their coefficients have not been thoroughly investigated. % the characteristics can be very different from conventional image data.
Thus, we present a compression pipeline for our purposes.
To automatically filter out unnecessary coefficients, we propose a trainable binary mask.
For each 3D scene, we jointly optimize grid parameters and their corresponding masks.
This per-scene optimization strategy can be more optimal than the global quantization table used in standard image compression codecs~\cite{jpeg}.

% encoding methods...
To compress sparse grid representations, we first merge masks with wavelet coefficients to zero out coefficients and then apply standard compression algorithms to masked coefficients.
We utilize the run-length encoding to encode binary information about which coefficients are non-zero. % the information about which coefficients are zeros, stored in the learned masks.
For further compression, we apply one of the entropy coding algorithms, Huffman encoding~\cite{huffman}, to these encoded outputs. % these encoded outputs are fed into an entropy coding algorithm, such as Huffman encoding~\cite{huffman}. % , for further compression.

Our method incurs negligible computational costs at test time, requiring only one inverse DWT (IDWT) per grid.
After the IDWT, the wavelet grids are transformed into spatial grids, eliminating the need for additional IDWT during rendering.
As a result, the computing time and costs (including memory costs) are identical to the original spatial grid-based NeRF.

In summary, our contributions are as follows:
\begin{itemize}
    \item We propose using wavelet coefficients to improve parameter sparsity and reconstruction quality. Through experiments, we show that the wavelet coefficients can be more compact than the spatial domain coefficients in neural radiance fields. % We propose using wavelet coefficients to improve the parameter sparsity and reconstruction quality. Experimental results show that the wavelet coefficients are more compact than the spatial domain representation.
    \item We propose a trainable mask that can be applied to any grid-based neural representation. Experimental results demonstrate that our proposed masking method can zero out more than 95\% of the total grid parameters while maintaining high reconstruction quality. % We propose a novel masking method for a more compact 3D scene and object representation, which applies to general grid-based neural representations. Throughout the experiments, we show that we can mask out more than 95\% of the total grid parameters while maintaining high reconstruction quality.
    \item We achieve state-of-the-art performance in novel view synthesis under a memory budget of 2 MB.
\end{itemize}

%-------------------------------------------------------------------------

\section{Related Works}
\subsection{Neural rendering}

% Neural fields 전반적인 설명과 한계
\noindent\textbf{Neural radiance fields.} NeRF~\cite{nerf} has demonstrated that 3D scenes and objects can be successfully represented using coordinate-based neural representation.
NeRF renders a scene by casting a ray per pixel and sampling colors and opacity from points that lie on the ray.
However, since this approach relies on the dense, point-wise sampling of color values and opacity, the computational costs are significant.

% 이를 해결하기 위한 grid 기반...
Due to this inefficiency, there has been considerable interest in improving computational efficiency.
A plethora of methods propose to adopt data structures such as trees~\cite{PlenOctrees, Fourier-plenoctree}, point clouds~\cite{point-nerf,pointlightfields,zhang2022differentiable,Shape-as-points}, and grids~\cite{Plenoxels,snerg,instant-ngp,VQAD,tensorf,DVGO,schwarz2022voxgraf} to efficiently represent 3D scenes and objects.
PlenOctrees~\cite{PlenOctrees} use an octree structure for real-time rendering.
Despite the reduced rendering costs, the increased size of more than ten times the size of standard MLP-based neural fields, such as NeRF~\cite{nerf}, limits their wide application.
% Tree-based representations~\cite{PlenOctrees,Fourier-plenoctree} use an octree structure for real-time rendering, but the size of these octrees is much larger than that of standard MLP-based neural fields. % To achieve real-time rendering, tree-based representations~\cite{PlenOctrees} use an octree structure to represent 3D scenes and objects, but the size of these octrees exceeds two orders of magnitude that of NeRF.
Point cloud-based approaches~\cite{point-nerf,pointlightfields} reduce the number of points sampled per ray by utilizing the scene geometry of a point cloud.
Lastly, a number of studies have demonstrated the efficiency of using grid structures, including vectors, matrices, and tensors, on neural fields~\cite{Plenoxels,instant-ngp,tensorf,DVGO}.
With the use of these grid structures, training iterations that once required tens of hours can now be completed in a matter of minutes.
% Lastly, introducing grids has considerably reduced the training time~\cite{instant-ngp,tensorf,DVGO}, making it possible to finish training within a few minutes instead of tens of hours.
Furthermore, real-time rendering at inference time is also possible. % made possible.
However, dense 3D grid structures require substantial amounts of memory, often exceeding 1 GB.
To alleviate this memory requirement, several methods have been proposed to reduce the size of grids~\cite{instant-ngp,VQAD,tensorf,CCNeRF}.
% Nevertheless, since memory storage for grid coefficients is also significant, several methods have been proposed to reduce the size of grids~\cite{instant-ngp,VQAD,tensorf}.
% Grid-based approaches~\cite{instant-ngp,tensorf,DVGO} trade off a large memory footprint for a reduction in training and inference time by utilizing the grid structure.

To achieve near-instant rendering with reasonable memory requirements, Instant-NGP~\cite{instant-ngp} proposed using hash-based multi-resolution grids. Using hash functions, each grid maps input coordinates to corresponding feature vectors.
% To accelerate training and inference times and lower the computational costs of neural fields, Instant-NGP~\cite{instant-ngp} proposes using hash-based multiresolution grids where feature vectors for each coordinate can be extracted.
% cNeRF~\cite{cnerf} exploits entropy loss on neural field to compress the model weight by imposing additional loss term with respect to the estimated entropy.
Similarly, VQAD~\cite{VQAD} proposed using codebooks and vector quantization rather than the hash function.
This enables control over the overall size of the neural fields.
However, during training, a sizable amount of memory is required to learn which code from the codebook to use for each point on the grid.

In order to alleviate the memory requirement of using 3D grids, another line of work decomposes 3D grids into lower dimensional representations, such as planes and vectors~\cite{EG3D,tensorf,huang2022pref,CCNeRF}. % , to alleviate the memory requirement.
EG3D~\cite{EG3D} proposes a tri-plane approach, which represents a 3D scene with three perpendicular planes and uses feature vectors, extracted separately from each plane, as inputs for the following MLPs.
TensoRF~\cite{tensorf} is another parameter-efficient yet expressive method for decomposing dense three-dimensional grids into a smaller set of parameters, such as planes and lines.
Although these methods dramatically reduce the time and space complexity of the 3D scene and object representation, their overall sizes are larger than those of MLP-only methods. % they still require a lot more memory storage than basic MLP-based methods.

% K-plane 언급?

% 이외에 frequency 기반...
\noindent\textbf{Frequency-based representation.} % Standard signal compression methods exploit the compactness of signal representation in the frequency domain.
Several studies have explored frequency-based parameterization for efficient neural field representations~\cite{Fourier-plenoctree,huang2022pref,filterbank,tancik2020fourier}. % For efficient representation in neural fields, several works have explored frequency-based parameterization.
Fourier PlenOctree~\cite{Fourier-plenoctree} has demonstrated that the Fourier transform can improve both parameter efficiency and training speed of the PlenOctree structure~\cite{PlenOctrees}.
% The training time has been significantly reduced, and the parameter efficiency has also improved.
Nevertheless, the overall size is larger than MLP-only methods.
PREF~\cite{huang2022pref} has also exploited the Fourier transform in 3D scene representation.
%Another attempt to use the Fourier transform in 3D scene representation is PREF~\cite{huang2022pref}.
It applies the Fourier transform to a tensor decomposition-based representation.
However, it has not shown notable improvements in both the representation quality and the parameter efficiency compared to spatial grid representations.
%Both the representation quality and the parameter efficiency, however, did not improve compared to spatial grid representations.
% It applies the Fourier transform to decomposed grids, but the overall performance (both the representation quality and the size) still falls behind the spatial grid representations.
% Fourier Plenoctree~\cite{Fourier-plenoctree} applied Fourier transform (FT) on Plenoctree~\cite{PlenOctrees} and PREF~\cite{huang2022pref} proposed a 3D phasor volume where frequencies are distributed uniformly along a 2D plane and dilated along a 1D axis.
% Complex-valued FT, however, is inefficient for compact neural representation because it requires twice the number of bits due to complex numbers.
% To achieve the same size as other approaches, the number of channels would have to be cut in half.
% \textcolor{red}{However, FT is not the optimal option for neural representations in that} % why..

% Discrete cosine transform(DCT) represents a signal with respect to a sum of cosine functions with various frequencies. 
% DCT tends to have most of its energy in a few coefficients, which in turn leads to the energy compaction. % need background?
% With its property of energy compaction, it has been widely used for image compression~\cite{}. 
% Specifically, DCT in image compression is employed in block-wise manner, typically $8 \times 8$ blocks, which incurs block artifact as interblock correlation is not considered~\cite{}. Additionally, DCT suffers more from visual distortion in low bit rates compared to medium bit rates~\cite{}.

Besides the Fourier transform, there is another line of signal representations in the frequency domain: the wavelet transform.
The wavelet transform decomposes a signal using a set of basis functions called wavelets.
The wavelet transform can extract both frequency and temporal information, in contrast to the Fourier transform, which can only capture frequency information. % Instead of focusing solely on global features as FT does, DWT can also analyze local features by using spatial locality in multiple resolutions.
Because each coefficient covers a different frequency and period in time, it is known that the wavelet transform can represent transient signals more compactly than Fourier-based transforms.
% The multi-resolution coefficients in regard to several frequency levels enable the compact representation of the original frame.
This compactness contributes to standard compression codecs, such as JPEG 2000~\cite{jpeg2000}. % , which has demonstrated the effectiveness of DWT coefficient compression with a narrowed bit width.
Motivated by the success of the standard codecs, we aim to mimic their compression pipeline and apply it to the neural radiance fields.
One disadvantage of the wavelet transform is that it can be less efficient than Fourier-based transforms for smooth signals.
However, we believe that most situations in grid-based neural fields are not the case, considering that each grid's resolution is constrained to accurately depict detailed scenes and objects with limited memory budgets.

\subsection{Model compression}
In recent years, many works have studied to downsize neural networks through various techniques such as weight pruning and quantization.

\textbf{Pruning.}
For efficient representations, pruning methods have been explored in neural fields.
Point-NeRF~\cite{point-nerf} proposes a pruning strategy for point cloud representation by imposing a sparsity loss on point confidence.
Re:NeRF~\cite{renerf} has validated that simply applying a general pruning technique, which gradually removes coefficients throughout the course of training, to grid-based neural representations results in a significant performance drop.
It proposed iterative parameter removal and conditional parameter inclusion and demonstrated comparable compression performance. However, it has not shown desirable performance under limited memory budgets.
Similarly, KiloNeRF~\cite{kilonerf} prunes out parameters for unused areas using a fixed threshold.
%In order to prune out coefficients, these methods rely on a fixed threshold.
Instead, we propose a trainable mask that can remove a large number of coefficients while causing only minor degradation in the reconstruction quality.

\textbf{Quantization.}
% cNeRF~\cite{cnerf} exploits entropy loss on neural field to compress the model weight by imposing additional loss term with respect to the estimated entropy.
% TODO: add ReNERF and VQAD
In terms of neural network quantization, there are two main approaches: post-quantization (PTQ)~\cite{brecq,pmlr-v119-nagel20a} and quantization-aware training (QAT)~\cite{NIPS2016_d8330f85,binaryconnect,zhu2017trained,xnor-net}.
Post-quantization methods quantize network parameters after training is finished.
PTQ has the advantage that it does not require additional training iterations or datasets.
Furthermore, it supports arbitrary-bit quantization. % can quantize neural networks for an arbitrary bit of precision.
Performance, however, might degrade because the network parameters were not optimized for the target bit precision.
Unlike post-quantization, QAT optimizes neural networks for a certain bit precision during training iterations.
Since quantization is not differentiable, it usually relies on the straight-through estimator (STE)~\cite{straight-through-estimator} for network optimization~\cite{binaryconnect}.
QAT has been validated for its effectiveness in neural representation models.
However, PTQ has the disadvantage of requiring additional training samples and iterations when quantizing the weights of a pretrained network. % A drawback of QAT is that quantizing a network that has already been trained requires training samples and additional training iterations for parameter quantization.
Since we train neural fields from scratch, we use PTQ for weight quantization.

\begin{figure*}[t]
\begin{center}
\includegraphics[width=1\linewidth]{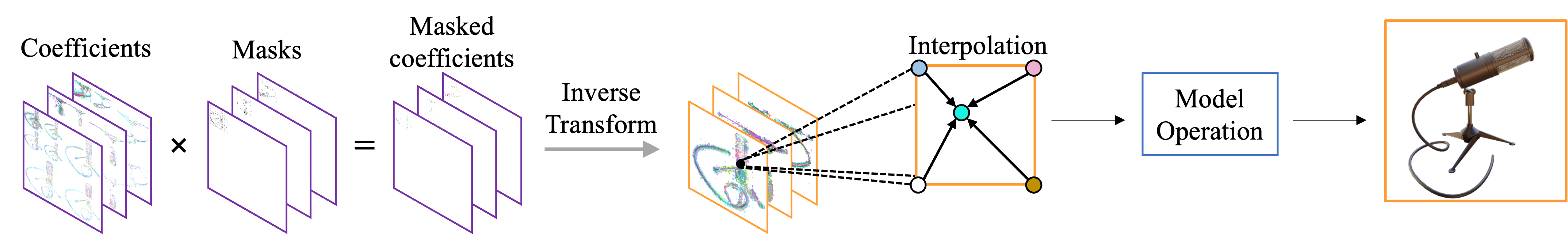}
\end{center}
   \caption{The overall architecture of our model. Images with purple borders illustrate wavelet coefficients, while those with orange borders visualize spatial coefficients.
   % White and orange blocks in 1D grids indicate masked features and unmasked features respectively.
   The wavelet coefficients in each 2D grid are multiplied with a binarized mask to form a masked wavelet coefficient grid. Masked wavelet coefficients are then inverse-transformed to spatial features. 
   % 1D spatial grids are also multiplied by their corresponding binarized masks.
   % We sample feature vectors for input coordinates using bilinear (linear) interpolation on 2D (1D) grids.
   We sample feature vectors for input coordinates using bilinear interpolation on the grids.
   Then, opacity and color for each input coordinate are estimated using the sampled feature vectors following the defined model.
   % We multiply and add feature vectors from 1D and 2D grids in order to estimate opacity. 
   % For color estimation, sampled feature vectors are concatenated (denoted with $\oplus$) and then fed to the following MLP, which processes feature vectors to estimate corresponding colors.
   }
\label{fig:pipeline}
\end{figure*}

\section{Methods}

\subsection{Neural radiance fields}
\label{ssec:nerf}
We consider a neural radiance field leveraging grid representations.
It takes an input coordinate $x \in \mathbb{R}^3$ and viewing direction, $d \in \mathbb{R}^2$, generating a four-dimensional vector consisting of a density and three-channel RGB colors, defined as follows.
\begin{align}
    &\sigma(x) = f_\sigma(x; \gamma_\sigma, \mathcal{M}_\sigma), \\
    &c(x,d) = f_c(x,d; \theta, \gamma_c, \mathcal{M}_c),
\end{align}
where $\theta$ is the parameter of an MLP, $\gamma=\{\gamma_\sigma, \gamma_c\}$ is a set of grid parameters, and $\mathcal{M}=\{\mathcal{M}_\sigma, \mathcal{M}_c\}$ is a set of masks for grid parameters, which will be described shortly (\cref{ssec:masking}).
The following volumetric rendering equation is used to synthesize novel views: % Given the neural radiance fields, volumetric rendering is typically used to synthesize novel views.
\begin{align}
    C(r) = \int_{t_n}^{t_f} T(t)\sigma(r(t))c(r(t),d)dt, \\
    T(t) = \text{exp}(-\int_{t_n}^{t} \sigma(r(s))ds), 
\end{align}
where $r(t)$ is a ray from a camera viewpoint, and $C(r)$ is the expected color of the ray $r(t)$. Two integral bounds (near and far) are denoted as $t_n$ and $t_f$, respectively. % where $r(t)$ is a ray from a camera viewpoint, and we compute the expected color $C(r)$ by integrating from near to far bounds $t_n$ and $t_f$, respectively.
Please refer to NeRF~\cite{nerf} for more details.

\subsection{Wavelet transform on decomposed tensors}
\label{ssec:dwt}
% A naive approach for compressing 3D grid representation is to apply classical frequency-based algorithms, such as DCT or DWT.
Frequency-based algorithms, such as discrete cosine transform (DCT) or discrete wavelet transform (DWT), have been developed and improved over the decades.
For high compression performance, standard codecs have used them to make sparse representations. %, and the standard codecs have utilized them due to their great compression performance.
However, in the case of 3D grid representation, their computational complexity increases cubically with grid resolution, making high-resolution 3D scene and object representations impractical.
In order to use frequency-based algorithms for compact representation, we need lower-dimensional data, such as 2D planes or 1D lines.

Recent studies have explored various decomposition methods to reduce the spatial complexity of 3D grid-based representations.
Among them, plane-based representations have achieved remarkable success in reducing the number of parameters while maintaining the rendering performance~\cite{tensorf,EG3D,huang2022pref}.
They propose decomposing a 3D tensor into a set of 2D planes or 1D vectors. % both 2D planes and vectors together.

To combine the best of both worlds, we propose using the wavelet transform on 2D plane-based neural fields~\cite{tensorf}. % combining 2D DWT with a 2D plane-based tensor decomposition~\cite{tensorf}.
We use the wavelet transform because of its compactness, especially for non-repeating, non-smooth signals.

For efficient 3D object and scene representation, recent studies proposed using a lower-dimensional grid (2D planes or 1D lines)~\cite{EG3D,tensorf}.
% Various combinations of 2D planes and operations can produce 3D grid representation $\gamma$.
For example, EG3D~\cite{EG3D} utilizes three 2D planes (tri-plane) and TensoRF~\cite{tensorf} employs three sets, each consisting of a plane and a vector. % for 3D representation.
We describe our method based on TensoRF.
% In this work, we use a set of 2D matrices and 1D vectors for grid representation,
In detail, we use a set of 2D matrices and 1D vectors for grid representation,
$\gamma_\sigma=\{\mathcal{W}_{\sigma,r}^{x}, \mathcal{W}_{\sigma,r}^{y}, \mathcal{W}_{\sigma,r}^{z}, v_{\sigma,r}^x, v_{\sigma,r}^y, v_{\sigma,r}^z\}_{r=1}^{N_{\sigma,r}}$ ($\sigma$ denotes density, and we will omit the subscript $\sigma$ for brevity from now on) as proposed in TensoRF~\cite{tensorf}.
% As proposed in TensoRF~\cite{tensorf}, we use a set of 2D matrices and 1D vectors for grid representation, % we define the parameters of grid representation as decomposed 2D matrices and vectors,
% $\gamma_\sigma=\{\mathcal{W}_{\sigma,r}^{x}, \mathcal{W}_{\sigma,r}^{y}, \mathcal{W}_{\sigma,r}^{z}, v_{\sigma,r}^x, v_{\sigma,r}^y, v_{\sigma,r}^z\}_{r=1}^{N_{\sigma,r}}$ ($\sigma$ denotes density, and we will omit the subscript $\sigma$ for brevity from now on).
$N_r$ is the number of ranks in matrix-vector decomposition and $\mathcal{W}_r^x \in \mathbb{R}^{H \times W}$, $\mathcal{W}_r^y \in \mathbb{R}^{W \times D}$, $\mathcal{W}_r^z \in \mathbb{R}^{H \times D}$ are matrices, $v_r^x \in \mathbb{R}^D$, $v_r^y \in \mathbb{R}^H$, $v_r^z \in \mathbb{R}^W$, are vectors in $x,y,z$ directions, respectively.
$H,W,D$ are the resolution of the grid.
% Note that we only apply DWT over matrices, therefore, $\mathcal{W}$ represents the coefficients of DWT, and $v$ contains information in the spatial domain.
More formally, a 3D grid representation $G$ can be defined as follows.
\begin{equation}
    % G =\sum_{r=1}^{N_r} v_r^{x} \otimes \texttt{idwt}(\mathcal{W}_r^{x}) + v_r^{y} \otimes \texttt{idwt}(\mathcal{W}_r^{y}) + v_r^{z} \otimes \texttt{idwt}(\mathcal{W}_r^{z}),
    G =\sum_{r=1}^{N_r}\sum_{d \in \{x, y, z\}} v_r^{d} \otimes \texttt{idwt}(\mathcal{W}_r^{d}),
    \label{eq:grid-representation}
\end{equation}
where $\otimes$ denotes the outer product and $\texttt{idwt}(\cdot)$ is a two-dimensional inverse discrete wavelet transform (IDWT).

With a slight abuse of notation, a single level IDWT of a matrix $\mathcal{W} \in \mathcal{R}^{m \times n}$ can be written as follows.
\begin{equation}
\begin{aligned}
    \texttt{idwt}(\mathcal{W}) = \Phi \star (\uparrow2)(\mathcal{W}_{LL}) + \Psi_{HL} \star (\uparrow2)(\mathcal{W}_{HL}) \\
    + \Psi_{LH} \star (\uparrow2)(\mathcal{W}_{LH}) + \Psi_{HH} \star (\uparrow2)(\mathcal{W}_{HH}),
\end{aligned}
\end{equation}
where $\mathcal{W}_{LL},\mathcal{W}_{HL},\mathcal{W}_{LH},\mathcal{W}_{HH} \in \mathbb{R}^{m/2 \times n/2}$ denote approximation, horizontal, vertical, and diagonal coefficients, respectively.
$\star$ denotes 2D cross correlation and $(\uparrow 2): \mathbb{R}^{m/2 \times n/2} \rightarrow \mathbb{R}^{m \times n}$ denotes (nearest neighbor) upscaling.
% where $\mathcal{W}_{LL} \in \mathbb{R}^{H/2 \times D/2}$, $\mathcal{W}_{HL} \in \mathbb{R}^{H/2 \times D/2}$, $\mathcal{W}_{LH} \in \mathbb{R}^{H/2 \times D/2}$, and $\mathcal{W}_{HH} \in \mathbb{R}^{H/2 \times D/2}$ denote approximation, horizontal, vertical, and diagonal coefficients.
% $\star$ denotes 2D cross correlation and $(\uparrow 2): \mathbb{R}^{H/2 \times D/2} \rightarrow \mathbb{R}^{H \times D}$ denotes (nearest neighbor) upscaling.
$\Psi$ and $\Phi$ are wavelet function and scaling function, respectively.
We use the biorthogonal 4.4 wavelet function and its corresponding scaling function~\cite{cohen1992biorthogonal}.
Even though we follow the grid representation of TensoRF, our method is not limited to the existing 2D plane-based representations but can be applied to any other approaches that adopt 2D planes to represent higher-dimensional structures.
% Although we follow the grid representation as TensoRF presented but it is worth to note that the proposed method can be applied to any grid representation which can be formed with the numerous combinations of 2D planes.

As shown in Eq.~\ref{eq:grid-representation}, we only apply IDWT over matrices $\mathcal{W}$ since we observed that IDWT to vectors resulted in significant quality degradation.
As this transformation is differentiable, the model can be trained end-to-end.
We can also obtain the 3D grid representation for appearance through the same process and provide all formulations in the supplementary material. % We can apply the same method to the appearance grid and provide all formulations in the supplementary material due to the page limits.

Our method incurs insignificant costs during inference.
Once training is finished, only one IDWT per grid is needed to transform wavelet grids into spatial grids.
Thus, the remaining computational costs and time are exactly the same as the original spatial grid-based neural fields.
% Despite the fact that wavelet coefficients must be transformed into spatial coefficients before being used, the costs of using the wavelet transform for inference are negligible.
% A one-time inverse transformation suffice to convert wavelet coefficients into spatial coefficients.
% The remaining computation costs and time are the same as for spatial grids.
% Before inference, each grid only requires a single inverse transformation.
% After that, the computation costs and time are the same as for spatial grids. % During inference, we only require a single inverse transformation for each plane, and after that, the remaining computational costs and time are the same as for spatial grids.

\subsubsection{Multi-level wavelet transform}
\label{sssec:mldwt}
\begin{figure}[t]
\begin{center}
\includegraphics[width=0.9\linewidth]{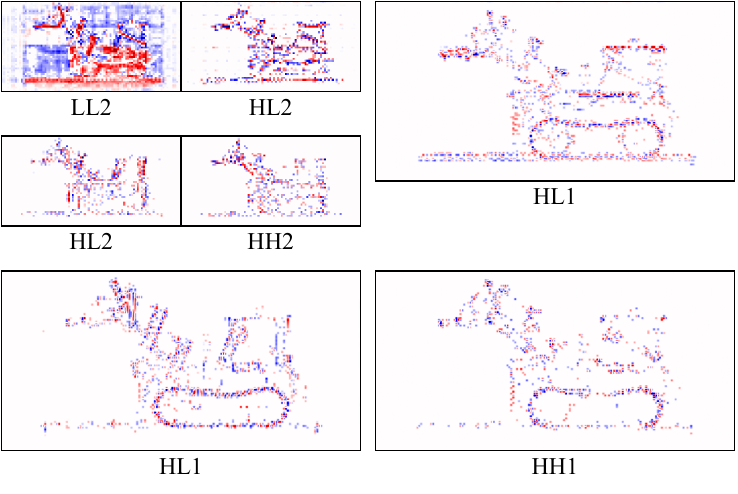}
\end{center}
   \caption{Multi-level discrete wavelet coefficients (2-level). LL, HL, LH, and HH denote approximation, horizontal, vertical, and diagonal coefficients, respectively. The numbers on the right denote the level of decomposition.}
\label{fig:dwt}
\end{figure}

To further improve parameter sparsity and reconstruction quality, we propose using a multi-level wavelet transform.
We experimentally found that higher-level wavelet transformations result in higher sparsity in grids (\cref{ssec:dwt_level}).
However, naively using a multi-level wavelet transform on grid parameters degrades the representation quality (\cref{tab:ablation}). % However, if the multi-level wavelet transform is applied carelessly, the quality could degrade.
We hypothesize that this is because the range of wavelet coefficients and the range of their gradients vary depending on the level of decomposition. %, which correspond to different decomposition levels, varies.
% Frequency-based representations of neural fields appear to suffer from varying scales of gradients and weights depending on locations, in contrast to spatial grids that have similar gradient and weight scales regardless of their locations.
In more detail, high-pass coefficients typically have a larger gradient scale with a smaller weight range.
Different weight scales and gradients for different frequencies seem to hinder the optimization process. 
As such, we propose multiplying wavelet coefficients by the scales proportional to the inverse of frequency.
With our proposed scaling factors $s$, Eq.~\ref{eq:grid-representation} can be rewritten as follows.
\begin{equation}
    G =\sum_{r=1}^{N_r}\sum_{d \in \{x, y, z\}} v_r^{d} \otimes \texttt{idwt}(\mathcal{W}_r^{d} \odot s_r^d),
    \label{eq:new-grid-representation}
\end{equation}
where $\odot$ denotes the Hadamard product.
For example, a two-level DWT, as shown in \cref{fig:dwt}, generates a total of seven groups of coefficients (LL2, HL2, LH2, HH2, HL1, LH1, HH1).
The scaling factors $s$ for HL2, LH2, and HH2 will be set to 1/2, and for HL1, LH1, and HH1 to 1/3.
% Among these, we multiply HL2, LH2, and HH2 by 1/2, and HL1, LH1, and HH1 by 1/3.
This scaling method enhances the quality of the reconstruction; more experiments are described in \cref{ssec:dwt_weight}.

\subsection{Learning to mask}
\label{ssec:masking}
Even though Wavelet coefficients can be sparse, we need additional methods to attain higher sparsity. % Since DWT can make representations sparse, we can exploit low entropy by utilizing various compression methods to reduce the overall size.
Thus, we propose using element-wise masks to increase the portion of zero elements in grids.
By jointly optimizing binary masks and grid parameters, we aim to zero out the majority of coefficients without significantly degrading the rendering quality.
As we defined earlier, we have a set of trainable element-wise masks, $\mathcal{M}=\{ \mathcal{M}_r^{x}, \mathcal{M}_r^{y}, \mathcal{M}_r^{z}, m_r^x, m_r^y, m_r^z\}_{r=1}^{N_r}$ (note that we omitted the subscript $\sigma$ for brevity).
During training, the grid parameters and corresponding binarized masks are multiplied element by element.
% We perform the element-wise multiplication between the grid parameters and the binarized masks.
Since calculating gradients directly from binarized masks is not feasible, we used the straight-through-estimator technique~\cite{straight-through-estimator} to train and use masks.
Formally, the masked matrix parameters $\widehat{\mathcal{W}}_r$ for $\mathcal{W}_r$ can be expressed as follows,
\begin{align}
    \widehat{\mathcal{W}}_r = \texttt{sg}((\mathcal{H}(\mathcal{M}_r) - \sigma(\mathcal{M}_r)) \odot \mathcal{W}_r) + \sigma(\mathcal{M}_r) \odot \mathcal{W}_r,
\end{align}
where $\texttt{sg}(\cdot)$ is the stop-gradient operator. % and $\odot$ denotes the Hadamard product. % is an element-wise product.
$\sigma(\cdot)$ and $\mathcal{H}(\cdot)$ denote the element-wise sigmoid and Heaviside step function, respectively.
We can also compute the masked vector parameters $\widehat{v}_r$ similarly. % Similarly, we can compute the masked parameters for all in $\mathcal{M}$, including masks for vectors $m$.
By replacing grid parameters ($\mathcal{W}_r$, $v_r$) from \cref{eq:new-grid-representation} with masked grid parameters ($\widehat{\mathcal{W}}_r$, $\widehat{v}_r$), we can represent a masked 3D grid representation as follows: % Then, a 3D grid with the masked parameters can be written as
\begin{equation}
    % \widehat{G} =\sum_{r=1}^{N_r} \widehat{v}_r^{x} \otimes \texttt{idwt}(\widehat{\mathcal{W}}_r^{x}) + \widehat{v}_r^{y} \otimes \texttt{idwt}(\widehat{\mathcal{W}}_r^{y}) + \widehat{v}_r^{z} \otimes \texttt{idwt}(\widehat{\mathcal{W}}_r^{z}).
    G =\sum_{r=1}^{N_r}\sum_{d \in \{x, y, z\}} \widehat{v}_r^{d} \otimes \texttt{idwt}(\widehat{\mathcal{W}}_r^{d} \odot s_r^d)
\end{equation}
To make binarized mask values sparse, we use the sum of all mask values as the additional loss term $\mathcal{L}_m$. The overall loss function is a sum of rendering loss $\mathcal{L}_r$ and the mask regularization term $\mathcal{L}_m$.
\begin{align}
    \mathcal{L} = \mathcal{L}_r + \lambda_m \mathcal{L}_m.
    \label{eq:mask_weight}
\end{align}
We use $\lambda_m$ to control the sparsity of the parameters of the grid representation.

\begin{figure}[t]
\begin{center}
\includegraphics[width=1.0\linewidth]{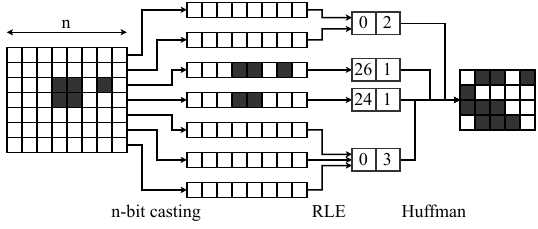}
\end{center}
   \caption{Proposed bitmap compression pipeline. RLE and Huffman denote run-length encoding and Huffman encoding, respectively.}
\label{fig:comp}
\end{figure}

\subsection{Compression pipeline} % Mask compression}
\label{ssec:encoding}
With our proposed masking method and multi-level DWT, the ratio of zeros in grids can go up to more than 95\%.
However, sparse representations themselves do not reduce the total size.
In this section, we describe our proposed compression pipeline for sparse grid parameters.
Instead of storing grids as they are, we separately store non-zero coefficients and bitmaps (or masks) that indicate which coefficients are non-zero.
Despite using 1-bit bitmaps, the overall bitmap size is large due to the large number of parameters.
To reduce the bitmap sizes, we propose a compression pipeline with the following three stages: n-bit casting, run-length encoding (RLE), and the Huffman encoding~\cite{huffman}.
The compression pipeline is illustrated in \cref{fig:comp}.

Before applying the compression pipeline, we split wavelet coefficient masks by level of decomposition. % As the wavelet coefficients corresponding to the same levels show similar levels of sparsity, we first separate the coefficients by the level.
For the 2-level wavelet transform, for instance, we group HL1, LH1, and HH1, then HL2, LH2, and HH2, and finally LL2 (\cref{fig:dwt}).
This is based on our experimental results that wavelet coefficients with the same levels of sparsity have similar levels of sparsity (\cref{fig:qual_mask}).
We found that grouping coefficients with similar sparsity results in better compression performance.
In addition, we found that directly applying RLE to 1-bit streams causes inefficient bit allocation due to the numerous repeating zeros.
Therefore, we first cast the binary mask values to n-bit unsigned integers, and perform RLE afterward.
In our experiments, we set n to 8.
Finally, we apply the Huffman encoding algorithm to the RLE-encoded streams to map values with a high probability to shorter bits.

\begin{figure*}[t]
\begin{center}
\includegraphics[width=0.95\linewidth]{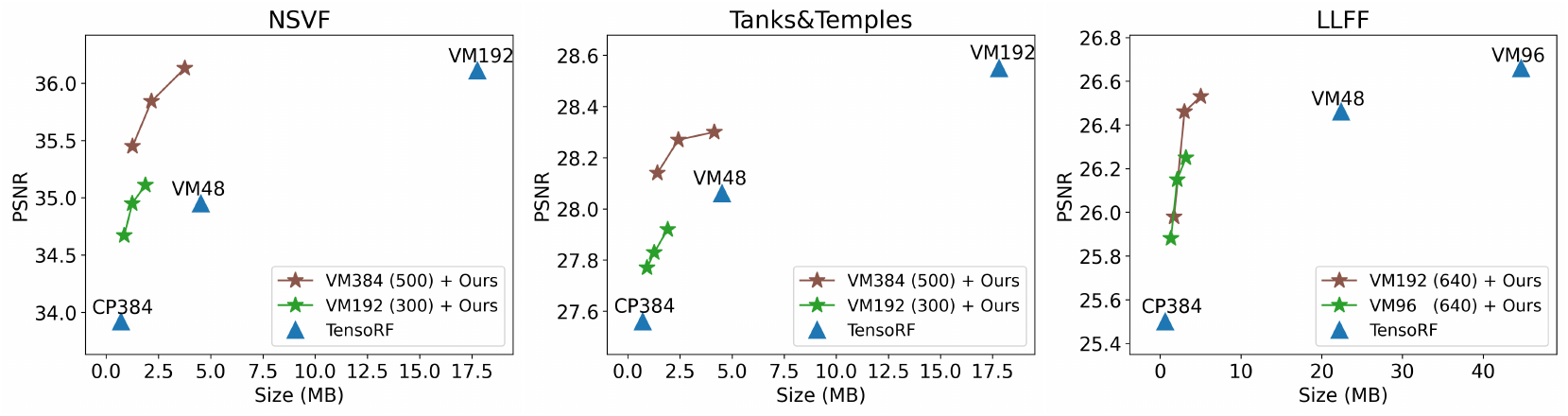}
\end{center}
\vspace{-0.5cm}
   \caption{Rate distortion curves on the NSVF synthetic, TanksAndTemples, and LLFF dataset. We used 8-bit precision for every method.}
\label{fig:rate_distortion_large_multi}
% \vspace{-0.5cm}
\end{figure*}

\section{Experiments}
\label{sec:experiments}

\subsection{Experimental settings}
We compared our proposed method with other baselines in four datasets; NeRF synthetic dataset~\cite{nerf}, Neural Sparse Voxel Fields (NSVF)~\cite{NSVF} dataset, TanksAndTemples dataset~\cite{tanksandtemples}, and LLFF dataset~\cite{mildenhall2019llff}.

We used TensoRF VM-192 as the baseline model and applied the proposed method.
% We used TensoRF-VM-192 as the baseline model and applied the proposed method.
To control the size of our method, we adjusted the mask regularization weight $\lambda_m$, the number of grid channels (from VM-192 to VM-384), and the resolution of grids.
% The mask regularization weight $\lambda_m$ controls the size of our method.
Unless otherwise specified, we used 8-bit quantization for every method. % For quantization, we used 8-bit grid coefficients.
We compared our method with quantized NeRF, TensoRF models (CP and VM), cNeRF~\cite{cnerf} and CCNeRF~\cite{CCNeRF}.
We also tried to compare our approach with VQAD~\cite{VQAD}, which efficiently compresses 3D scene representations using codebooks and vector quantization.
However, VQAD requires depth maps to prune out vacant areas in advance.
% However, we excluded VQAD because it requires depth map of training images to construct octree in which indices to appropriate codes in codebook are stored, while our experimental datasets does not provide depth map information and our model can be trained without it.
% More importantly, the required memory size grows exponentially with the target bit precision at inference time.
Otherwise, the required memory exceeds the memory capacity of the Tesla A100 equipped with 40 GB of memory.
Since the datasets we used do not provide depth maps, we did not compare VQAD with ours.
% Therefore, we cannot optimize VQAD on the datasets we used for evaluation.

We followed the experimental settings of TensoRF~\cite{tensorf}.
We trained models for 30,000 iterations, each of which is a minibatch of 4,096 rays.
We used the Adam optimizer and an exponential learning rate decay scheduler.
Following TensoRF, the initial learning rates of grid-related parameters and MLP-related parameters were set to 0.02 and 0.001, respectively.
Final learning rates were set at 1/10 of the initial learning rate.
We updated the alpha masks at the 2000th, 4000th, 6000th, 11000th, 16000th, 21000th, and 26000th iterations.
We set $\lambda_m$ in \cref{eq:mask_weight} to 1e-10 for high parameter sparsity and 5e-11 and 2.5e-11 for relatively lower sparsity.
We set the initial values of masks $\mathcal{M}$ to one, and set their initial learning rate to the same learning rate as grid parameters $\phi$.
As a reconstruction quality measurement, we used PSNR.

% The size of the models can be controlled by either changing the value of $\lambda_m$ or other grid-relevant hyperparameters such as resolution and channel.
% To enlarge our models beyond 2MB for the further experiments, we raised the resolution of grid from 300 to 500 and doubled the number of channels of grid to 384.
% Our enlarged models with only increased resolution, those with only doubled channels, and those with both increased resolution and double channels are dubbed with VM-192 (300) + Ours, VM-192 (500) + Ours, and VM-384 (500) + Ours, respectively.

\begin{figure}
\centering
\begin{subfigure}[b]{0.49\linewidth}
    \centering
    \includegraphics[width=\linewidth]{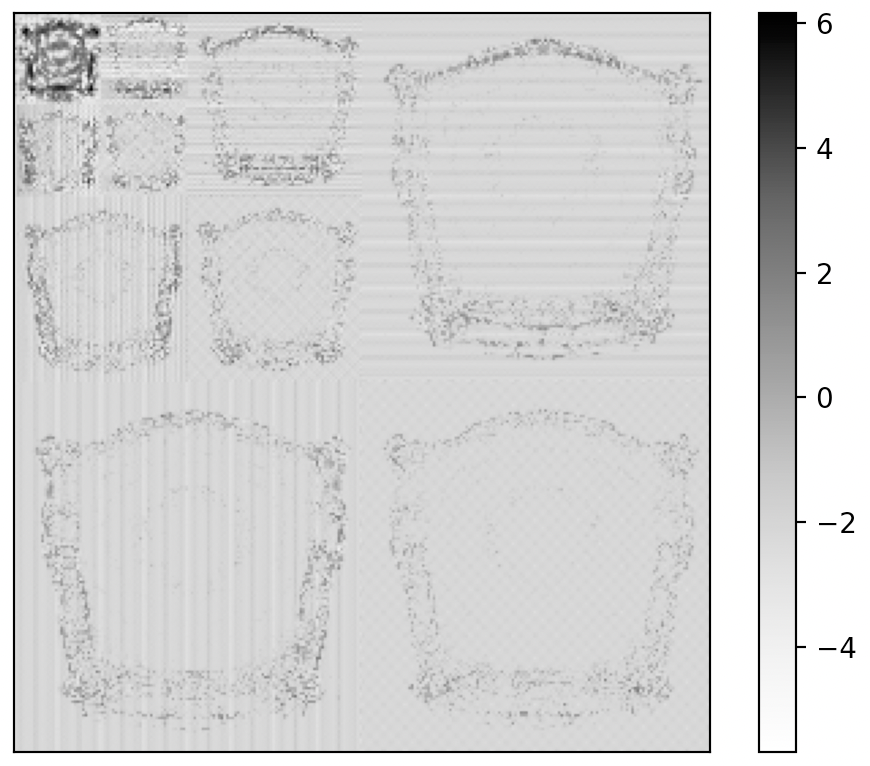}
    \caption{Raw mask}
    \label{fig:qual_mask_raw}
\end{subfigure}
\begin{subfigure}[b]{0.41\linewidth}
    \centering
    \includegraphics[width=\linewidth]{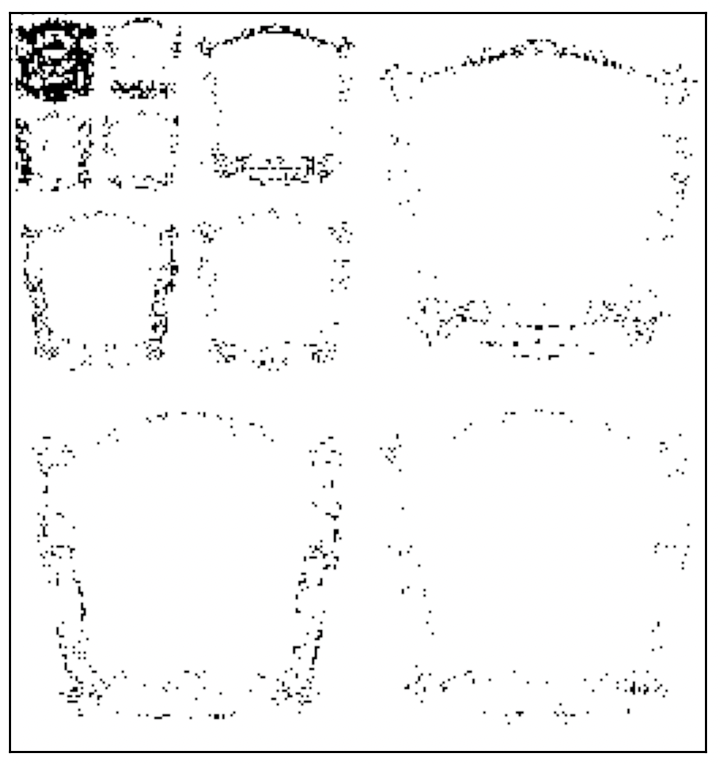}
    \caption{Binarized mask}
    \label{fig:qual_mask_binary}
\end{subfigure}
\caption{Qualitative analysis on mask $\mathcal{M}$ (Chair in the NeRF synthetic dataset).}
% \vspace{-0.8cm}
\label{fig:qual_mask}
\end{figure}

\subsection{Results}
\label{ssec:results}

\cref{fig:main-rate-distortion,fig:rate_distortion_large_multi} show the quantitative performances of our method and the baselines on various datasets.
Each graph displays the average PSNR and size of methods.
More detailed numbers are available in the supplementary materials.

As the rate-distortion curves show, our method achieves higher efficiency than the baseline models and other methods.
Even with the base network (VM-192), our method consistently outperforms other baselines on all novel view synthesis datasets, under a small memory budget of 2 MB.
We adjusted both $\lambda_m$ and the network size to increase the output size.
By adjusting $lambda_m$, we can control both size and quality while having no effect on computational costs, time, or memory requirements.
Increasing network size (doubling the number of channels and increasing the resolution) can push the boundary even further.
It is also worth mentioning that applying our method to a larger backbone model (VM-384) outperforms models with a smaller backbone model (VM-192), even with the same value of $\lambda_m$ or a similar model size.
However, as the network grows in size, the computational costs and time required for training and testing increase.
Therefore, it is still reasonable to use a small backbone model when using our proposed method.

\begin{figure*}[t]
\begin{center}
\centering
\includegraphics[width=0.9\linewidth]{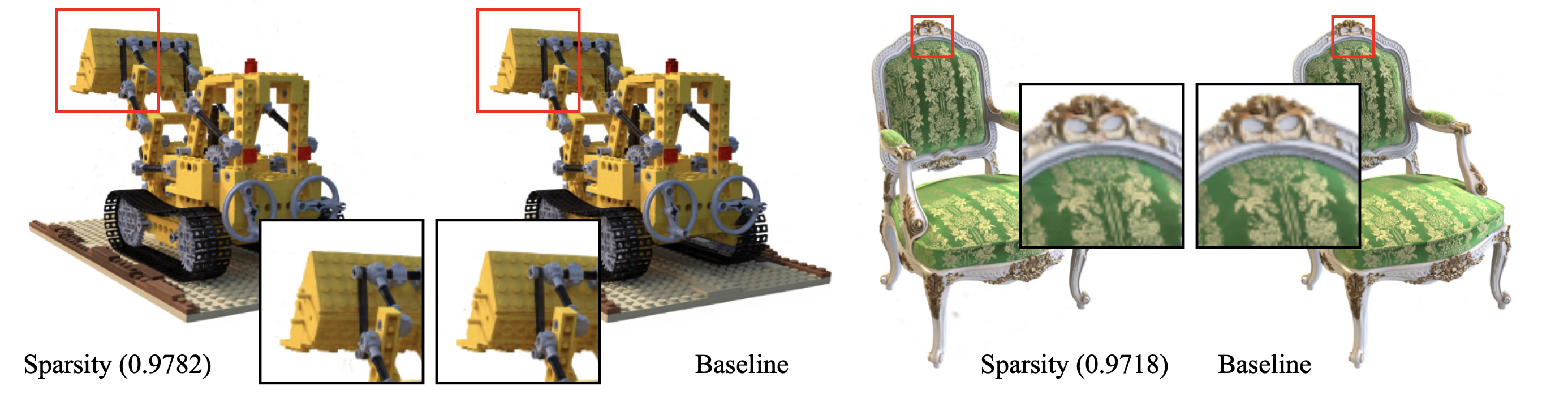}
\end{center}
   \caption{A qualitative analysis of grid sparsity. The sparsity refers to the ratio of zero elements in grids.
   With the help of trainable masks, removing more than 97\% of grid parameters only results in an imperceptible difference.}
\label{fig:qual}
\end{figure*}

\cref{fig:qual} shows the qualitative results with and without our proposed masking method.
Trainable masks removed more than 97\% of the grid parameters; nevertheless, the rendered results are still accurate, and the qualitative difference is almost imperceptible. % Although more than 97\% of the grid parameters are removed by the proposed masking approach, the rendered results are still satisfactory and the qualitative difference is almost visually imperceptible.
This demonstrates that our proposed method efficiently eliminates unnecessary wavelet coefficients by leveraging trainable masks.

\cref{fig:qual_mask} illustrates both raw and binarized masks after training.
First of all, sparsity varies depending on the level of the wavelet transform.
As shown in \cref{fig:qual_mask_binary}, coefficients with lower frequencies have lower sparsity, whereas coefficients with higher frequencies have higher sparsity.
What is interesting is that the raw mask values seem to reflect the characteristics of corresponding wavelet coefficients.
Vertical, horizontal, and diagonal patterns can be found in the raw mask values for the vertical, horizontal, and diagonal coefficients, respectively. % The raw mask values for vertical features contain vertical patterns; raw mask values for horizontal features contain horizontal patterns; and raw mask values for diagonal features contain diagonal patterns.
% We believe these patterns demonstrate that mask values are well optimized with corresponding wavelet coefficients.

\subsection{Ablation studies}
In this section, we analyze each component of our proposed method.
Every experiment was conducted on the whole NeRF synthetic dataset, and we used the average PSNR as a measurement for representation quality. % , averaging the PSNR of each scene.
We used 4-level wavelet transform and set $\lambda_m$ to 1e-10, unless otherwise specified.

\begin{figure}[t]
\begin{center}
\includegraphics[width=1.0\linewidth]{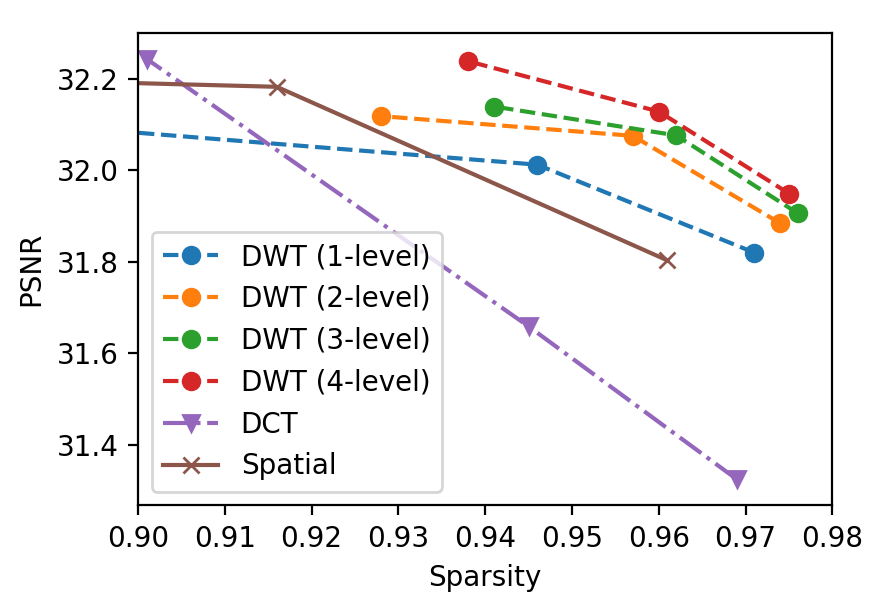}
\end{center}
\vspace{-0.25cm}
   \caption{The rate distortion curves of different signal representation schemes (spatial, DWT, and DCT). % The effect of the level of DWT on grid sparsity and reconstruction quality.
   Sparsity on the x axis refers to the ratio of zeros in grid parameters. % parameters in the grids.
   The grid sparsity was controlled by $\lambda_m$.
   % These are average performances on the NeRF synthetic dataset.
   The numbers inside the parenthesis indicate the levels of the wavelet transform. %  level of the wavelet transform is indicated inside the parenthesis.
   }
% \vspace{-1cm}
\label{fig:dwt_level}
\end{figure}

\subsubsection{Discrete wavelet transform}
\label{ssec:dwt_level}
In this section, we evaluate the compactness of wavelet coefficients and the performance improvements brought by the multi-level wavelet transform. % In this section, we analyze the compactness of wavelet coefficients and how the level of multi-level DWT affects reconstruction performance. % as measured by PSNR.
\cref{fig:dwt_level} shows the rate-distortion curves of four different levels of wavelet transform and non-wavelet coefficients. % the curves of four different levels of wavelet decomposition compared to those without wavelet decomposition.
As shown in the figure, wavelet-based grid representation outperforms the grid in the spatial domain, especially when the sparsity is high.
Furthermore, rate-distortion curves demonstrate that higher level of wavelet transform improves visual quality and sparsity. % the higher the level of DWT, the better the visual quality it gets with much higher sparsity.

\subsubsection{Level-wise wavelet coefficient scaling}
\label{ssec:dwt_weight}
As mentioned in \cref{sssec:mldwt}, we proposed multiplying wavelet coefficients by the scales proportional to the inverse of frequency.
Thus, we compared two versions of wavelet transform; one with our proposed weight scaling (\cref{sssec:mldwt}) and the other without it.
As \cref{tab:ablation} shows, scaling wavelet coefficients improves reconstruction performance.

\begin{table}[t]
    \centering
    \begin{tabular}{c|c}
    \hline
    Methods & PSNR \\
    \hline\hline
    \textbf{DWT} & \textbf{31.949} \\
    DWT w/o level-wise scaling & 31.145 \\
    DCT & 31.325 \\
    \hline
    \end{tabular}
    \caption{Ablation study on level-wise scaling.
    Without our proposed scaling method, the performance of DWT is worse than DCT.
    % When using the proposed level-wise weight, DWT outperforms DCT in terms of reconstruction quality as measured in PSNR.
    }
    \label{tab:ablation}
\end{table}

\begin{table}[t]
    \centering
    \begin{tabular}{c||c}
    \hline
    Wavelet Function & PSNR \\
    \hline\hline
    Haar & 31.889 \\
    Coiflets 1 & 31.846 \\
    Daubechies 4 & 31.734 \\
    \textbf{biorthogonal 4.4 (default)} & \textbf{31.949} \\
    reverse biorthogonal 4.4 & 31.727 \\
    \hline
    \end{tabular}
    \caption{Reconstruction performances of different wavelet functions.}
    \label{tab:wav_func}
\end{table}

\subsubsection{Wavelet functions}
In this section, we analyze how different wavelet functions affect reconstruction quality.
For comparison, we selected several wavelet functions: Haar, Coiflet 1, Daubechies 4, and reverse biorthogonal 4.4.
As shown in \cref{tab:wav_func}, the type of wavelet function has little effect on reconstruction quality.
Even Haar, the simplest wavelet function, performs fairly well.
Still, the biorthogonal 4.4 that we used shows the best performance among selected wavelet functions.

\subsubsection{Discrete cosine transform}
In this section, we compare the performance of DWT with that of DCT, which represents a signal with respect to the sum of cosine functions with different frequencies.
DCT, similar to DWT, has an energy compaction property.
That is, a small number of non-zero DCT coefficients are sufficient to represent a given signal. % Similar to DWT, when a signal is converted with DCT, non-zero values tend to concentrate in a small number of bins, which in turn leads to energy compaction.
Because of its energy compactness, DCT or its variants are another widely used signal transformation for image, audio, and audio compression.
However, as shown in \cref{fig:dwt_level} and \cref{tab:ablation}, DWT shows superior compression and representation performance to DCT. % replacing DWT with DCT degrades reconstruction quality.
% Unlike signal understanding in the frequency domain, optimizing DCT coefficients appears to be demanding.
We believe this is due to the non-repeating, non-smooth information in grids, for which DWT is more appropriate.

\section{Conclusion}
We propose a compact representation for grid-based neural fields, enabled by a novel masking strategy and multi-level wavelet transform. % on plane-based neural fields.
We demonstrate that these two components made it possible to remove more than 95\% of grid parameters without a significant loss of visual quality. % while preserving reconstruction quality.
% Without DWT, we discovered that the performance loss brought on by parameter sparsity is rather significant.
With our proposed compression pipeline and the sparse wavelet coefficients, we achieved state-of-the-art performance under a memory budget of 2 MB.
We believe we can reduce the size further with more sophisticated compression pipelines.
% Still, we believe that there is still room for further compression through elaborating the compression pipeline.

Our proposed method is naturally constrained by the limitations of grid-based neural fields that are developed for bounded scenes or objects.
We believe that expanding this grid-based representation to encompass unbounded or large scenes would be an intriguing direction, given that we can now compactly represent 3D objects with our proposed representation scheme.

\section*{Acknowledgments}
This work was supported in part by the Institute of Information and Communications Technology Planning and Evaluation (IITP) grants (2021-0-02068, IITP-2021-0-02052, IITP-2023-2020-0-01821, IITP-2019-0-00421) and National Research Foundation of Korea (NRF) grants (2022R1A4A3032913, 2022R1F1A1064184, 2022R1A4A3033571) funded by the Korea Government (MSIT).
%%%%%%%%% REFERENCES
{\small
\bibliographystyle{ieee_fullname}
\bibliography{egbib}
}

% \end{document}

% \newpage
\clearpage
\appendix
\renewcommand\thesection{\Alph{section}}
\setcounter{section}{0}

% \title{Masked Wavelet Representation for Compact Neural Radiance Fields: Supplementary Materials}
% \maketitle

\twocolumn[
\begin{center}
\Large{\bf{Masked Wavelet Representation for Compact Neural Radiance Fields: \\ Supplementary Materials}}\par\vspace{3ex}
\end{center}]

\section{Qualitative Comparison}
\begin{figure}[t]
\centering
\begin{center}
\includegraphics[width=\linewidth]{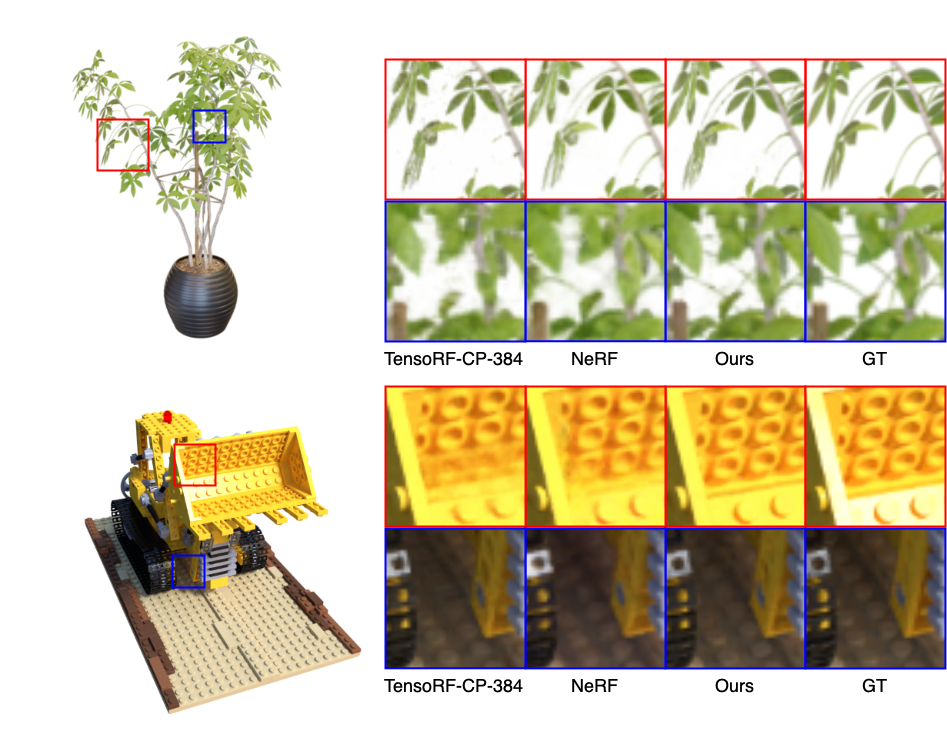}
\end{center}
\caption{Qualitative analysis of similar-sized representations. Every method was quantized to 8 bits and less than 1 MB.}
\label{fig:qual-sim-size}
\end{figure}

In this section, we visually demonstrate the qualities of neural fields with similar sizes. \cref{fig:qual-sim-size} shows the qualitative results of similar-sized representations: TensoRF-CP-384, NeRF, and Ours ($\lambda_m$=1e-10). The bit precision of each model was set to 8 bits. Even though they have similar overall sizes (less than 1 MB), ours shows the best qualitative results.

% (NeRF, CP, ours-small) : ficus, lego
% video: ours-large, VM-192(32)

\section{Computational Costs}
\label{sec:comp-cost}
We also present the computational costs, particularly in terms of time, because we introduced learnable masks and DWT for efficient scene representation.
First of all, as we explained in the main paper, we can ignore the computational costs at inference time.
% This is because once an inverse DWT (IDWT) is performed on the grids, the spatial grids are constructed and they are used for rendering.
% Thus, we do not need IDWTs nor masks during the test time and we can still enjoy the identical computing time and costs (including memory costs) to the spatial grid-based NeRF.
However, those two components incur non-negligible training time during training time.
\cref{tab:train-time} shows the approximate training time to train a network (VM-192).
It was evaluated using a 40 GB memory-equipped Tesla A100.
Introducing masks increases the training time, but by a relatively insignificant amount of time.
On the other hand, DWT increases training time more significantly.
We believe that the increased training time is due to two factors: additional computational costs caused by DWT and not fully optimized DWT codes.
As a result, we believe future DWT code optimization can help close this training time gap, especially given that current codes lower the GPU utilization rate from 80\% (baseline) to 50\% (4-level DWT with mask).
Nevertheless, the time required by our proposed method is still far less than that of neural fields that exclusively use MLP, such NeRF.

\begin{table}[t]
    \centering
    \begin{tabular}{c|c|c}
    \hline
     & No Mask & Mask \\
     \hline
    Spatial & $\approx$ 8 min & $\approx$ 9 min \\
    1-level DWT & $\approx$ 13 min & $\approx$ 14 min \\
    2-level DWT & $\approx$ 15 min & $\approx$ 17 min \\
    3-level DWT & $\approx$ 18 min & $\approx$ 20 min \\
    4-level DWT & $\approx$ 23 min & $\approx$ 24 min \\
    \hline
    \end{tabular}
    \caption{Approximate training time}
    \label{tab:train-time}
\end{table}

% Default: 8 minutes, 80\% (8.0)
% + mask: 8 minutes, 80\% (9.5)
% + 1-level DWT: 13 minutes 70\% (14.5) - no mask (13.)
% + 2-level DWT: 14 minutes 30 sec 60\% (17.5) - (15)
% + 3-level DWT: 18 minutes 60\% (20.5) - (18.5)
% + 4-level DWT: 21 minutes 50 \% (24.5) - (23)

\section{Comparison on Masking Method}
\begin{table}[]
    \centering
    \begin{tabular}{c|c|c}
        \toprule
        Methods & Sparsity & PSNR \\
        \hline
        Abs. Thres. ($\tau$=0.5) & 0.8554 & 32.33 \\
        % Abs. Thres. ($\tau$=0.9) & 0.9279 & 30.09 \\
        Abs. Thres. ($\tau$=1.0) & 0.9385 & 28.95 \\
        % Abs. Thres. ($\tau$=1.1) & 0.9474 & 27.72 \\
        % Abs. Thres. ($\tau$=1.3) & 0.9561 & 25.17 \\
        Abs. Thres. ($\tau$=1.6) & 0.9747 & 21.40 \\
        \hline
        % Ours & $\lambda_m$ & 5e-11 & 0.9694 & 32.13 \\
        Ours ($\lambda_m$=2.5e-11) & 0.9376 & 32.24\\
        Ours ($\lambda_m$=1.0e-10) & 0.9769 & 31.95 \\
        \bottomrule
    \end{tabular}
    \caption{Pruning on NeRF Synthetic dataset. The threshold of the absolute value-based method (Abs. Thres.), is denoted as $\tau$.}
    \label{tab:coef-trunc}
    \vspace{-1em}
\end{table}

We also compared our masking method with a threshold-based pruning approach that removes grid coefficients whose absolute values are below a threshold $\tau$.
% We also compared our masking method with absolute value-based truncation manner that prunes out grid coefficients whose absolute values are lower than threshold $\tau$.
As \cref{tab:coef-trunc} shows, ours is more parameter-efficient than the threshold-based pruning method.
It further supports the effectiveness of our proposed masking method in improving efficiency.
% It further proves the crucial role of the proposed masking method in improving efficiency.

\section{Ablation Studies on Mask Compression}

% \begin{table*}[t]
%     \centering
%     \resizebox{\linewidth}{!}{
%         \begin{tabular}{c||c||c|c|c|c|c|c|c|c}
%         \hline
%             & Avg & Chair & Drums & Ficus & Hotdog & Lego & Materials & Mic & Ship \\
%         \hline
%         \hline
%             None & 16.55 & 15.65 & 16.04 & 16.56 & 17.94 & 15.89 & 18.96 & 15.23 & 16.12 \\
%             Run-length encoding & 1.19 (13.96) & 1.39 (11.28) & 1.58 (10.16) & 1.47 (11.27) & 0.70 (25.62) & 1.08 (14.72) & 1.46 (12.97) & 0.73 (20.94) & 1.08 (14.94) \\
%             Huffman encoding & 0.45 (36.94) & 0.52 (29.98) & 0.59 (27.05) & 0.55 (29.94) & 0.26 (67.94) & 0.41 (38.56) & 0.55 (34.59) & 0.28 (54.38) & 0.41 (39.13) \\
%         \hline
%             + 8bit casting & 0.33 (50.60) & 0.40 (39.03) & 0.42 (38.20) & 0.42 (39.42) & 0.18 (101.91) & 0.31 (51.91) & 0.39 (48.48) & 0.20 (76.14) & 0.30 (53.38) \\
%             + Level-wise encoding & 0.28 (59.31) & 0.36 (43.72) & 0.35 (45.83) & 0.35 (47.31) & 0.16 (111.41) & 0.26 (60.86) & 0.34 (55.75) & 0.16 (96.99) & 0.26 (63.22) \\
%         \hline
%         \end{tabular}
%     }
%     \caption{The mask size and compression ratio at each stage of the compression pipeline evaluated on the NeRF synthetic dataset. Each number is in megabytes, and the number inside parenthesis indicates the compression ratio.} % The mask size is given in megabytes and the number in parentheses indicates the compression ratio.}
%     \label{tab:compress-ablation}
% \end{table*}

\begin{table*}[t]
    \centering
    \resizebox{\linewidth}{!}{
        \begin{tabular}{cccc||c||c|c|c|c|c|c|c|c}
        \hline
            Lv. wise. & 8-bit. & RLE & Huffman & Avg & Chair & Drums & Ficus & Hotdog & Lego & Materials & Mic & Ship \\
        \hline
        \hline
            & & & & 2.07 & 1.96 & 2.01 & 2.07 & 2.24 & 1.99 & 2.37 & 1.91 & 2.02 \\
            & & $\surd$ & & 1.19 (1.75) & 1.39 (1.41) & 1.58 (1.27) & 1.47 (1.41) & 0.70 (3.21) & 1.08 (1.84) & 1.46 (1.62) & 0.73 (2.62) & 1.08 (1.87) \\
            & & $\surd$ & $\surd$ & 0.45 (4.62) & 0.52 (3.75) & 0.59 (3.38) & 0.55 (3.74) & 0.26 (8.49) & 0.41 (4.82) & 0.55 (4.32) & 0.28 (6.78) & 0.41 (4.89) \\
            & $\surd$ & $\surd$ & $\surd$  & 0.33 (6.24) & 0.40 (4.88) & 0.42 (4.41) & 0.42 (4.93) & 0.18 (12.71) & 0.31 (6.48) & 0.39 (6.05) & 0.20 (9.50) & 0.30 (6.68) \\
            $\surd$ & $\surd$ & $\surd$ & $\surd$ & 0.28 (7.39) & 0.36 (5.47) & 0.35 (5.60) & 0.35 (5.91) & 0.16 (13.92) & 0.26 (7.61) & 0.34 (6.97) & 0.16 (12.13) & 0.26 (7.88) \\
        \hline
        \end{tabular}
    }
    \caption{The mask size and compression ratio at each stage of the compression pipeline evaluated on the NeRF synthetic dataset. Each number is in megabytes, and the number inside parenthesis indicates the compression ratio. RLE and Huffman indicate run-length and Huffman encoding, respectively. Level-wise encoding and 8-bit casting are denoted as ``Lv. wise." and ``8-bit.". The first row shows the original binarized mask size.}
    \label{tab:compress-ablation}
\end{table*}

\begin{table*}[t]
    \centering
    \resizebox{\linewidth}{!}{
    \begin{tabular}{c|c||c||c|c|c|c|c|c|c|c}
    \hline
    Methods & size(MB) & Avg & Chair & Drums & Ficus & Hotdog & Lego & Materials & Mic & Ship \\
    \hline
    \hline
    Baseline & 14.41 & 28.52 & 31.83 & 22.81 & 23.06 & 35.50 & 29.67 & 27.31 & 30.87 & 27.10 \\
    \hline
    Ours ($\lambda_m$=1e-10) & 0.71 & 27.91 & 30.69 & 23.18 & 23.22 & 33.18 & 38.85 & 27.09 & 30.29 & 26.79 \\
    Ours ($\lambda_m$=5e-11) & 0.98 & 28.35 & 30.89 & 23.39 & 23.30 & 34.65 & 29.19 & 27.36 & 30.696 & 27.03 \\
    Ours ($\lambda_m$=2.5e-11) & 1.38 & 28.39 & 31.07 & 23.18 & 23.27 & 34.78 & 29.18 & 27.44 & 31.05 & 27.14 \\
    \hline
    \end{tabular}}
    \caption{The performance of our proposed method with the tri-planar architecture on the NeRF synthetic dataset. Reconstruction quality was measured in PSNR. By setting $\lambda_m$ to 5e-11, our proposed method compresses the baseline model 14.70 times without significant PSNR loss.}
    % Performance on the NeRF synthetic dataset measured in PSNR with tri-planar architecture. Our method can compress the baseline model 14.70 times without significant PSNR loss, using our model with $\lambda_m$=5e-11.}
    \label{tab:triplane-nerf-dataset}
\end{table*}

% \begin{table*}[t]
%     \centering
%     \resizebox{\linewidth}{!}{
%         \begin{tabular}{c||c|c||c|c|c|c|c|c|c|c|c|c|c|c|c|c|c|c}
%             \toprule
%             & \multicolumn{2}{c}{Avg} & \multicolumn{2}{c}{Chair} & \multicolumn{2}{c}{Drums} & \multicolumn{2}{c}{Ficus} 
%             & \multicolumn{2}{c}{Hotdog} & \multicolumn{2}{c}{Lego} & \multicolumn{2}{c}{Materials} & \multicolumn{2}{c}{Mic} & \multicolumn{2}{c}{Ship} \\
%             \toprule
%             & Density & Appearance & Density & Appearance & Density & Appearance & Density & Appearance 
%             & Density & Appearance & Density & Appearance & Density & Appearance & Density & Appearance & Density & Appearance \\
%         \end{tabular}
%     }
% \end{table*}

\begin{table}[t]
    \centering
    \resizebox{\columnwidth}{!}{
        \begin{tabular}{c|c|c|c|c|c|c}
            \hline
                & Plane & Lv. 1 & Lv. 2 & Lv. 3 & Lv. 4 (precinct) & Lv. 4 (upper-left) \\
            \hline
            \hline
                \multirow{3}{*}{Density} & YX & 0.99 & 0.97 & 0.91 & 0.76 & 0.33 \\
                                         & ZX & 0.99 & 0.96 & 0.91 & 0.82 & 0.36 \\
                                         & ZY & 0.99 & 0.97 & 0.93 & 0.84 & 0.43 \\
            \hline
                \multirow{3}{*}{Appearance} & YX & 0.97 & 0.94 & 0.90 & 0.80 & 0.61 \\
                                            & ZX & 0.98 & 0.96 & 0.93 & 0.87 & 0.63 \\
                                            & ZY & 0.99 & 0.98 & 0.95 & 0.91 & 0.73 \\
            \hline
        \end{tabular}
    }
    \caption{Sparsity table of 4-level DWT with learnable mask ($\lambda_m$=1e-10) on Chair (from the NeRF dataset).}
    \label{tab:average-sparsity}
\end{table}

\begin{table}[]
    \centering
    \begin{tabular}{c|cccc}
        \toprule
         & no scan & zigzag & morton & spiral \\
        \hline
        Size (MB) & 0.83 & 0.83 & 0.85 & 0.83 \\
        \bottomrule
    \end{tabular}
    \caption{Average size by scanning order on the NeRF synthetic dataset.}
    \label{tab:scanning-type}
\end{table}

In this section, we analyze our proposed compression pipeline in more detail by demonstrating the compression ratio at each stage and outlining the rationale for our design decisions.
\cref{tab:compress-ablation} shows the compression ratios in the NeRF dataset.
By incrementing each step, we demonstrate how each step contributes to the compression ratio.
Keep in mind that we do not compress non-zero coefficients; we only compress information regarding which coefficients are non-zero (mask).
We keep the non-zero coefficients that are not compressed as well as the compressed mask that shows where the non-zero components are located.
% We conduct ablation studies to evaluate the compression ratio (CR) at each stage of the proposed compression pipeline and explain the reason behind our design choices in more detail.
% We first run-length encode (RLE) and Huffman encode the binarized masks and evaluate the CR.
%  Then we measure how much the CR increases when using 8-bit casting before the RLE. 
% We do not measure the improvement in CR when using only the casting without any encoding afterwards. 
% Note that we do not measure how the casting itself affects the CR.
% This is because it only packs 8 bits into a 1 byte and thus does not affect the size of the masks.
% Note that we do not measure how the casting itself affects the CR, as it does not change the size of the mask with no encoding afterwards.
% Furthermore, we partition the masks by the DWT level, compress each level of masks through the proposed pipeline, and measure the final CR.

\textbf{Run-length encoding (RLE)}
The RLE is effective for compressing data with repeating numbers. Instead of encoding raw repeating numbers, RLE encodes repeating numbers as a pair of the number and its count. This is why we adopt RLE in our mask compression pipeline. Our proposed method zeros out most grid coefficients. More specifically, by adopting our proposed method, the sparsity of grids can go up to 90\% or even 99\%. As shown in the second row of \cref{tab:compress-ablation}, adopting RLE can reduce the mask size by a factor of 1.75, on average.
% Since the RLE converts contiguous same numbers into a single number and its count, it is suitable for compressing data with many repeating numbers.
% Our loss term forces the mask to be sparse, and thus the learned masks have many repeating zeros. 
% As shown in the second row of \cref{tab:compress-ablation}, the RLE reduces the mask size by a factor of 13 on average.
% In fact, we found more than 99\%, 97\%, 92\% of the values were zero in the Lv. 1, Lv. 2, and Lv. 3 of the appearance mask, respectively (\cref{tab:average-sparsity}).
% This explains why the RLE efficiently compress the masks at the first stage.

\textbf{Huffman encoding} As shown in the third row of \cref{tab:compress-ablation}, we can raise the compression ratio to 4.62 by including Huffman encoding in our compression pipeline.
We also compared adaptive arithmetic coding ~\cite{adaptive_arithmetic} with Huffman encoding but did not observed meaningful improvements.
As a result, we chose to use the less expensive Huffman encoding in our compression pipeline.
We believe, however, that further improvements could be made by incorporating advanced compression techniques into the proposed method.

% \textcolor{red}{The mask size is further reduced 36 times smaller by Huffman encoding (\cref{tab:compress-ablation}). BLABLA.}

\textbf{8-bit casting}
Packing the binarized mask values by 8 bits before applying RLE can further increase the compression ratio to 6.24 (\cref{tab:compress-ablation}).
This is because the casting makes the length of the RLE code much shorter.
For example, consider a thousand 0s.
Without casting, we can represent a thousand 0s with three (0, 255) and one (0, 235).
On the other hand, with the help of casting, we only need a pair of two numbers (0, 125).
We use this method assuming most elements are zeros, and as shown in the table, it can really improve compression ratio.
%\textcolor{red}{Table ?} also shows that the length of the RLE coe decreases.

\textbf{Level-wise encoding}
We discovered that sparsity highly depends on the level of DWT, as shown in \cref{tab:average-sparsity}. More specifically, high-pass coefficients have higher sparsity. Based on the findings, we separate the mask at each level and compress separately. For the last level of DWT, we treat the upper left part (LL) and the remaining three parts (LH, HL, HH, also known as precincts) separately, as the former has much fewer zeros compared to the latter. As shown in \cref{tab:compress-ablation}, adding the level-wise encoding into the pipeline improves the compression ratio even further, resulting in an average size of 0.28 mb, which is 7.39 times smaller than the size of the original mask.
% All the numbers discussed so far are the results of encoding the masks as a whole, regardless of the DWT level.
% However, we found that the number of zeros in the mask increases as the DWT level decreases (\cref{tab:average-sparsity}).
% In other words, the sparsity of each level is different.
% Based on the findings, we compress the mask separately for each level.
% We even divide the last level of mask into the upper left part and the remaining three parts (the precinct), as the former has much less zeros compared to the latter.
% As shown in \cref{tab:compress-ablation}, adding the level-wise encoding into the pipeline results in the final mask size of 0.28mb, which is 59 times smaller than the size of the original mask.

% \textcolor{red}{It is worth noting that improvement in CR by RLE is different across the objects, but improvement by other components is roughly the same (\cref{tab:compress-ablation}).}
% Specifically, by adopting Huffman encoding, 8-bit casting, and level-wise encoding, the CR increases by 2.7 times, 1.4 times, and 1.2 times, respectively.
% This suggests that the CR of the proposed method is dependent on the RLE.
% Since the length of the RLE code depends on how contiguous the same number is in an input stream, the CR can be further increased by adopting a scanning method suitable for each object.

\textbf{Scanning order}
Since scanning order before compression can affect the output size, we also tried different scanning orders and measure the output sizes.
Tab.~\ref{tab:scanning-type} shows the results on the NeRF synthetic dataset.
As shown, scanning order did not affect the performance noticeably.

% \section{Additional Experiments}
% \subsection{Application to Tri-Planar Architecture}
\section{Application to Other 2D Grid-based Neural Fields}
Even though we only showed results using a TensoRF~\cite{tensorf} model (VM-192), our proposed method is not confined to TensoRF.
Only TensoRF was used in the main paper because it is currently the most effective method for plane-based neural fields.

To demonstrate that our method can be used with any grid-based neural representation, we apply it to additional plane-based neural fields in this section.
We exclusively employed 2D grids, not 1D grids, as plane-based neural fields.
This was only intended to demonstrate that our suggested method can be used with other 2D grid-based methods in addition to TensoRF.
TriPlane from EG2D~\cite{EG3D} served as the model for this design, which uses only 2D planes.
The difference is that, like TensoRF, we separate the grids for density and appearance.
For implementation, we removed 1D grids from the baseline model we used in the main paper.
% In this section, we conducted an extra experiment of using 2D grids only to train neural field, to validate that our method can be applied to any grid architectures.
% The model for experiment has six 2D planes, where three grids contain the DWT coefficients for density estimation and other three grids have those for color estimation, with six corresponding 2D masks.
% The experimental settings are same to those provided in the main paper except that 1D grids are not exploited.
% Therefore, the points interpolated from the 2D grids are directly summed up to compute opacity values or concatenated and then fed to the following MLP to estimate colors.

As shown in \cref{tab:triplane-nerf-dataset}, our proposed method can be successfully applied to other 2D grid-based representations and remove most of the coefficients without causing considerable quality degradation.
When $\lambda_m$ was set to 5e-11, our proposed method reduced the size to less than 7\% of its original size with negligible quality drops (0.17 drops measured in PSNR).

\section{Color Estimation in Detail}
Following TensoRF~\cite{tensorf}, we use separate grids for density and appearance (color) estimation. In this section, we describe appearance grids in detail.
% As TensoRF~\cite{tensorf} suggested, we have separated grids for each density and color estimation.
% In this section, we describe the grids for color estimation in detail.
Similar to density grids, we use three 2D matrices and three vectors, $\phi_c=\{\mathcal{W}_{c,r}^{x}, \mathcal{W}_{c,r}^{y}, \mathcal{W}_{c,r}^{z}, v_{c,r}^x, v_{c,r}^y, v_{c,r}^z\}_{r=1}^{N_{c,r}}$ ($c$ denotes color, and we will omit the subscript $c$ for brevity from now on). 
% Similar to density grids, we represent DWT coefficients and spatial domain information for color values as the parameters of grid which is decomposed into three 2D matrices and three vectors, $\phi_c=\{\mathcal{W}_{c,r}^{x}, \mathcal{W}_{c,r}^{y}, \mathcal{W}_{c,r}^{z}, v_{c,r}^x, v_{c,r}^y, v_{c,r}^z\}_{r=1}^{N_{c,r}}$ ($c$ denotes color, and we will omit the subscript $c$ for brevity from now on). 
$N_r$ is the number of ranks in matrix-vector decomposition and $\mathcal{W}_r^x \in \mathbb{R}^{H \times W}$, $\mathcal{W}_r^y \in \mathbb{R}^{W \times D}$, $\mathcal{W}_r^z \in \mathbb{R}^{H \times D}$ are matrices, $v_r^x \in \mathbb{R}^D$, $v_r^y \in \mathbb{R}^H$, $v_r^z \in \mathbb{R}^W$, are vectors in $x,y,z$ directions, respectively. $H,W,D$ denote the resolution of the grid. 
We employ DWT only over matrices, just as we did over density grids.
Thus, $\mathcal{W}$ are wavelet coefficients, and $v$ are feature vectors in the spatial domain. % information in the spatial domain.

Density grids in TensoRF only generate a density value, while color grids generate a feature vector for each coordinate.
These feature vectors are forwarded to MLP to estimate the colors.
To extract a feature vector per coordinate, TensoRF uses additional vectors ($f^x$, $f^y$, $f^z$).
% In order to extract a feature vector, instead of a scalar, for each coordinate in a 3D grid.
% Unlike the density grids, color grids need extra $3N_r$ vectors whose dimension is equal to the dimension of the color feature channel.
% In our experimental setting, the color feature channel dimension, 128, is lower than that of the planes which is 300, so the rank of the extra vectors is lower as well.
% Therefore, we do not integrate them into the 2D matrices but use them as additional vectors~\cite{tensorf}, denoted by $f_r$ and outer product them to the corresponding inverse-transformed matrices.  
More formally, a 3D grid representation $G_c$ can be defined as follows.
    
% \begin{equation}
% \begin{split}
%     G = &\sum_{r=1}^{N_r} v_r^{x} \circ \texttt{iwt}(\mathcal{W}_r^{x}) \circ f_{3r} + v_r^{y} \circ \texttt{iwt}(\mathcal{W}_r^{y}) \circ f_{3r-1} \\
%     & + v_r^{z} \circ \texttt{iwt}(\mathcal{W}_r^{z}) \circ f_{3r-2},
% \end{split}
% \end{equation}

% \begin{equation}
% \begin{aligned}
%     G = &\sum_{r=1}^{N_r} v_r^{x} \circ \texttt{iwt}(\mathcal{W}_r^{x}) \circ f_{3r} + v_r^{y} \circ \texttt{iwt}(\mathcal{W}_r^{y}) \circ f_{3r-1} \\
%     & + v_r^{z} \circ \texttt{iwt}(\mathcal{W}_r^{z}) \circ f_{3r-2},
% \end{aligned}
% \end{equation}

\begin{equation}
    G_c = \sum_{r=1}^{N_r} \sum_{d \in \{x, y, z\}} v_r^{d} \otimes  \texttt{idwt}(\mathcal{W}_r^{d}) \otimes  f_r^{d},
\end{equation}

where $\otimes $ denotes the outer product and $\texttt{idwt}(\cdot)$ is a two-dimensional inverse discrete wavelet transform. % with the biorthogonal 4.4 wavelet function. % During inference, only a single inverse transformation is applied to each plane, and after that, the remaining computational costs and time are the same as for spatial grids.

% We compared the performance of our method measured with respect to PSNR with the baseline where DWT and our novel masking mechanism are not applied but grids contain the information of spatial features.
% \cref{tab:triplane-nerf-dataset} shows that our method can also be applied to tri-plane structure to effectively compress the grid-based neural fields, using only 6.8\% of memory, without considerable performance loss compared to the baseline model.

% to effectively represent the grid-based neural fields with 93.2% of memory 93.2%

% \subsection{Experimental Results on Real-World Scenes}
% In order to evaluate our method on unbounded real-world scenes, we conducted extra experiment on LLFF~\cite{nerf} dataset. We applied NDC transformation to the scenes following the experimental settings of TensoRF.
% \label{sec:unbounded-scenes}

\section{Per-Scene Results}
\begin{figure*}[t]
\begin{center}
\includegraphics[width=0.75\linewidth]{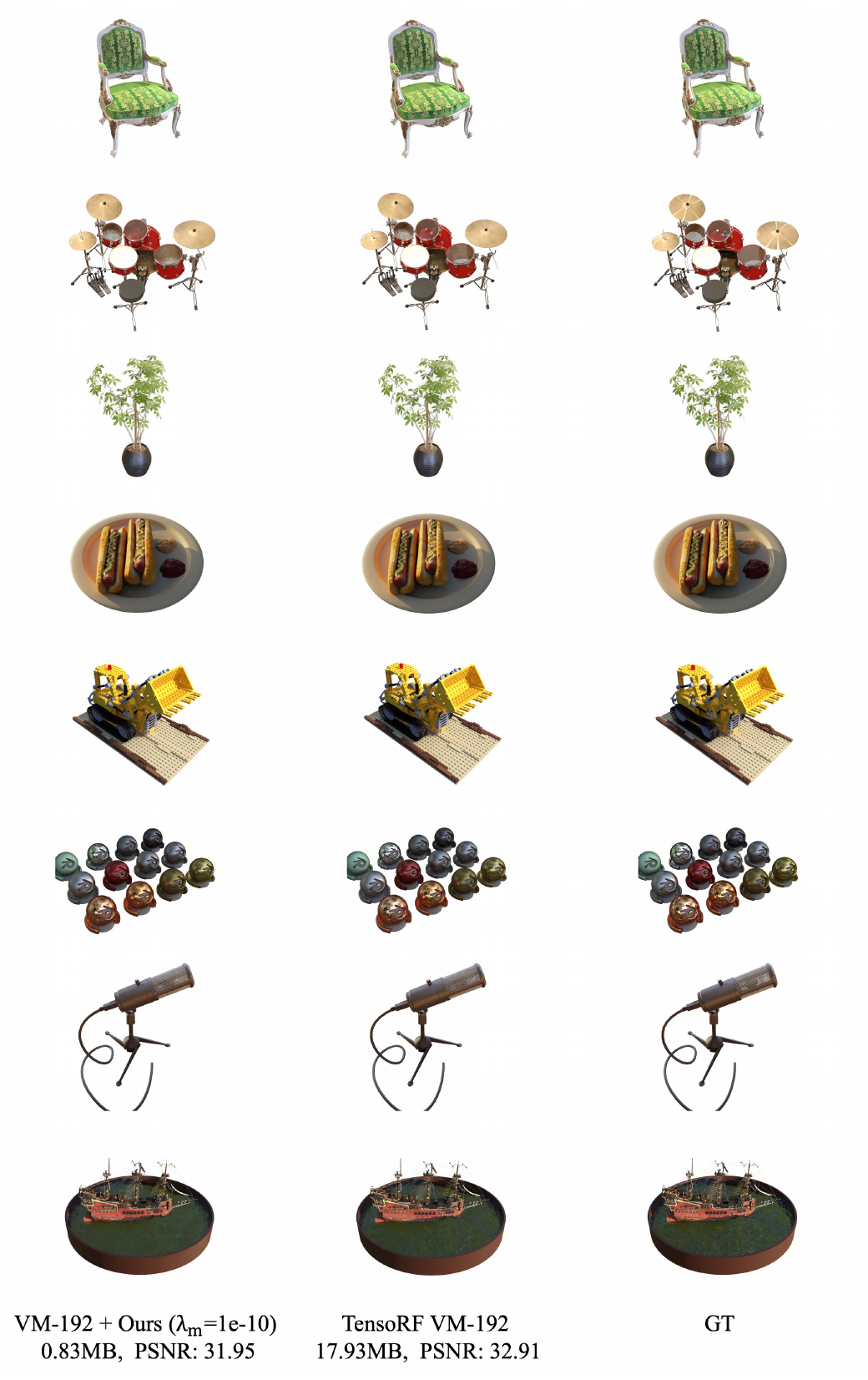}
\end{center}
\vspace{-0.5cm}
   \caption{Qualitative results on the NeRF synthetic dataset.}
\label{fig:NeRF-dataset}
% \vspace{-0.5cm}
\end{figure*}

\begin{figure*}[t]
\begin{center}
\includegraphics[width=0.75\linewidth]{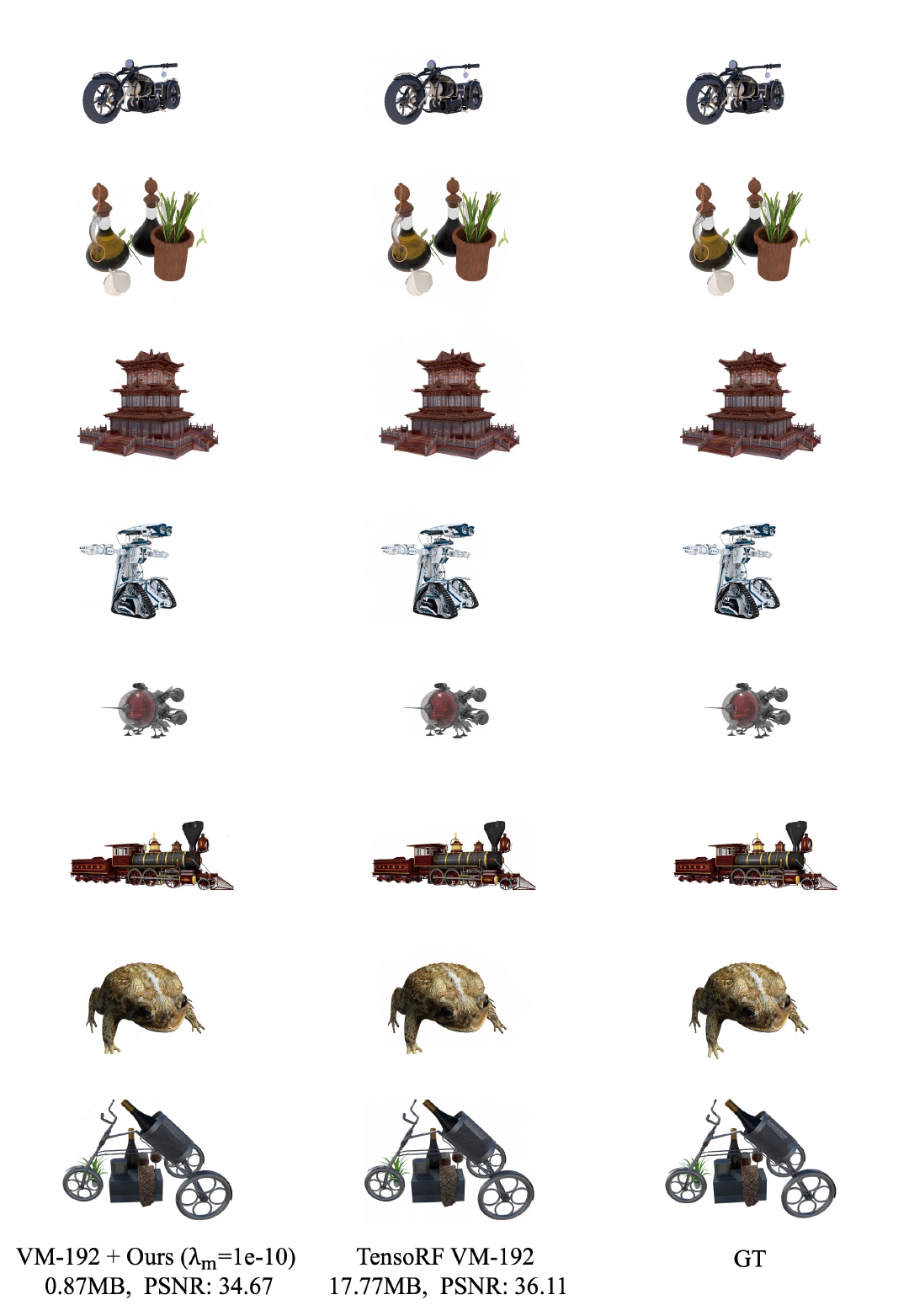}
\end{center}
\vspace{-0.5cm}
   \caption{Qualitative results on the NSVF dataset.}
\label{fig:NSVF-dataset}
% \vspace{-0.5cm}
\end{figure*}

\begin{figure*}[t]
\begin{center}
\includegraphics[width=0.8\linewidth]{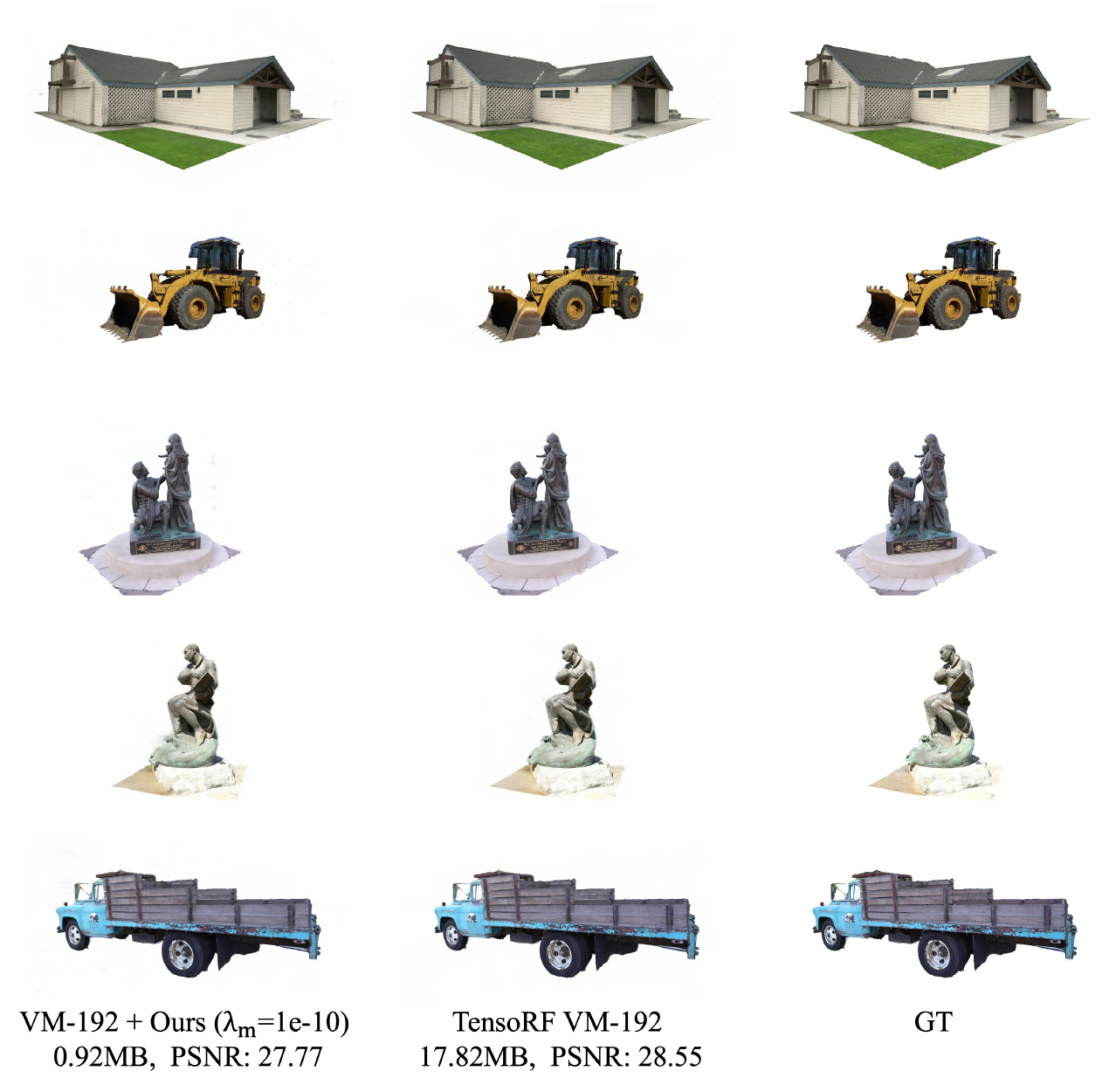}
\end{center}
\vspace{-0.5cm}
   \caption{Qualitative results on the Tanks\&Temples dataset.}
\label{fig:TNT-dataset}
% \vspace{-0.5cm}
\end{figure*}

\begin{figure*}[t]
\begin{center}
\includegraphics[width=0.75\linewidth]{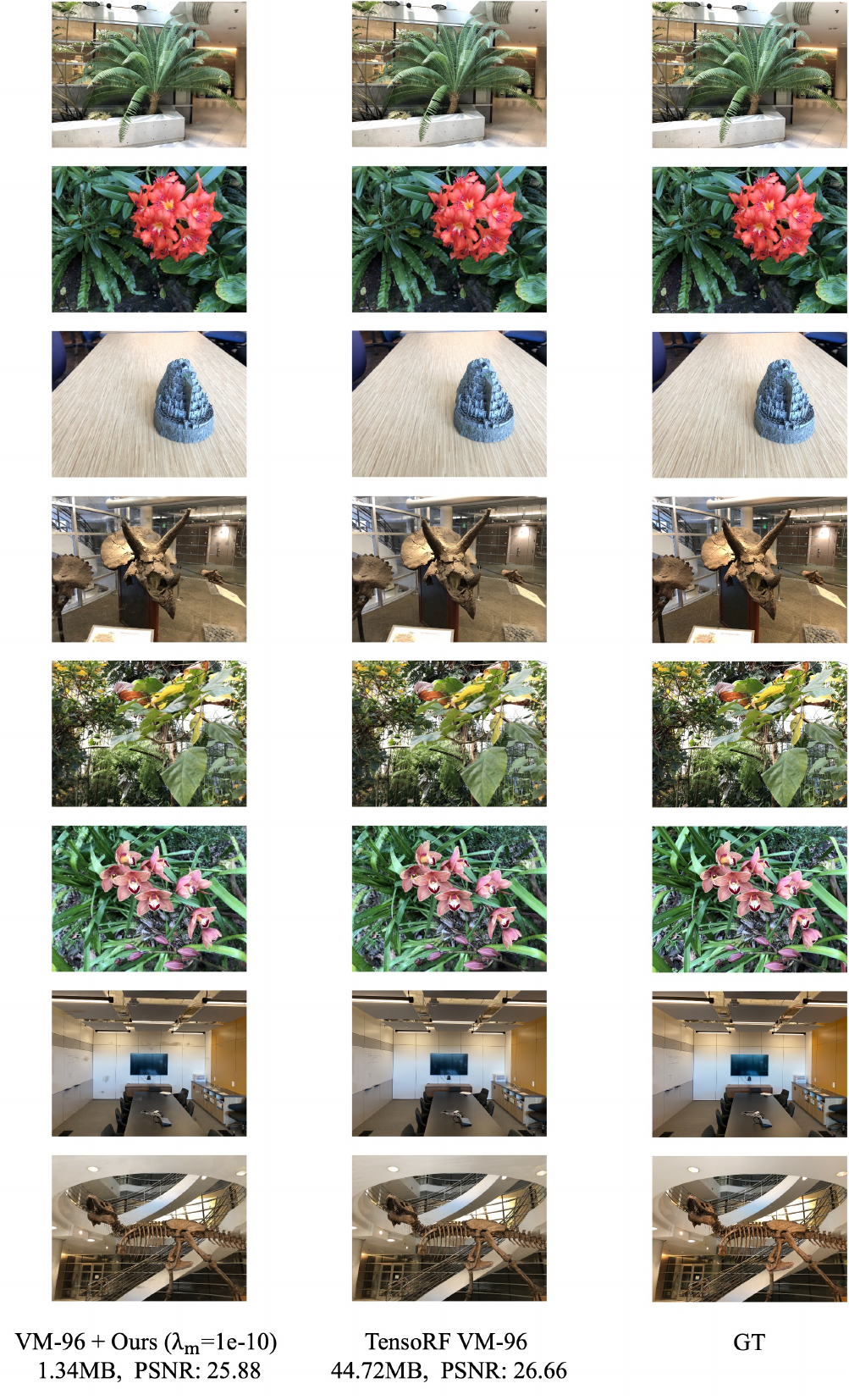}
\end{center}
\vspace{-0.5cm}
   \caption{Qualitative results on the LLFF dataset.}
\label{fig:LLFF-dataset}
% \vspace{-0.5cm}
\end{figure*}

In this section, we provide quantitative and qualitative results of each scene from NeRF synthetic, NSVF synthetic, TanksAndTemples, and LLFF dataset.
\cref{tab:nerf-dataset,tab:nsvf-dataset,tab:Tanks&Temples-dataset,tab:llff-dataset} show the quantitative results
and \cref{fig:NeRF-dataset,fig:NSVF-dataset,fig:TNT-dataset,fig:LLFF-dataset} show the qualitative results on the four datasets.

% \cref{tab:nerf-dataset-large} and \cref{fig:NeRF-dataset} shows quantitative and qualitative results on NeRF synthetic dataset. \cref{tab:NSVF-dataset-large} and \cref{fig:NSVF-dataset} shows quantitative and qualitative results on the NSVF synthetic dataset. \cref{tab:Tanks&Temples-dataset} and \cref{fig:Tanks&Temples-dataset} show the results on the Tanks\&Temples dataset. \cref{tab:llff-dataset} and \cref{fig:LLFF-dataset} show the results on the LLFF dataset.

\begin{table*}[t]
    \centering
    \resizebox{\linewidth}{!}{
    \begin{tabular}{c|c||c||c|c|c|c|c|c|c|c}
    \hline
    Methods & size(MB) & Avg & Chair & Drums & Ficus & Hotdog & Lego & Materials & Mic & Ship \\
    \hline
    \hline
    KiloNeRF $^\circ$ & $\leq$ 100 & 31.00 & 32.91 & 25.25 & 29.76 & 35.56 & 33.02 & 29.20 & 33.06 & 29.23 \\
    CCNeRF (CP) $^\circ$ & 4.4 & 30.55 & - & - & - & - & - & - & - & - \\
    NeRF $^*$ & 1.25 & 31.52 & 33.82 & 24.94 & 30.33 & 36.70 & 32.96 & 29.77 & 34.41 & 29.25 \\
    cNeRF $^\bullet$ & 0.70 & 30.49 & 32.28 & 24.85 & 30.58 & 34.95 & 31.98 & 29.17 & 32.21 & 28.24 \\
    \hline
    PREF $^*$ & 9.88 & 31.56 & 34.55 & 25.15 & 32.17 & 35.73 & 34.59 & 29.09 & 32.64 & 28.58 \\
    \hline
    VM-192 $^*$ & 17.93 & 32.91 & 35.64 & 25.98 & 33.57 & 37.26 & 36.04 & 29.87 & 34.33 & 30.64 \\
    VM-48 $^*$ & 4.81 & 32.18 & 34.54 & 25.55 & 33.12 & 36.60 & 35.14 & 29.15 & 33.33 & 29.99 \\
    CP-384 $^*$ & 0.72 & 31.18 & 33.49 & 25.11 & 29.86 & 35.97 & 33.26 & 29.56 & 33.56 & 28.59 \\
    \hline
    VM-192 (300) + Ours $^*$ & & & & & & & & & & \\
    $\lambda_m$=1e-10 & 0.83 & 31.95 & 34.14 & 25.53 & 32.87 & 36.08 & 34.93 & 29.42 & 33.48 & 29.15 \\
    $\lambda_m$=5e-11 & 1.16 & 32.13 & 34.52 & 25.66 & 33.03 & 36.20 & 35.16 & 29.58 & 33.68 & 29.19 \\
    $\lambda_m$=2.5e-11 & 1.69 & 32.24 & 34.68 & 25.56 & 33.17 & 36.37 & 35.50 & 29.56 & 33.74 & 29.34 \\
    % $\lambda_m$=1e-11 & 2.51 & 32.21 & 34.74 & 25.55 & 33.20 & 36.47 & 35.48 & 29.59 & 33.45 & 29.23 \\
    \hline
    VM-192 (500) + Ours $^*$ & & & & & & & & & & \\
    $\lambda_m$=1e-10 & 1.02 & 32.14 & 34.64 & 25.55 & 33.04 & 35.85 & 35.15 & 29.54 & 33.91 & 29.41 \\
    $\lambda_m$=5e-11 & 1.55 & 32.37 & 34.90 & 25.69 & 33.25 & 36.13 & 35.50 & 29.63 & 34.21 & 29.65 \\
    $\lambda_m$=2.5e-11 & 2.36 & 32.46 & 35.16 & 25.77 & 33.34 & 36.26 & 35.74 & 29.65 & 34.19 & 29.57 \\
    % $\lambda_m$=1e-11 & 3.99 & 32.52 & 35.04 & 25.57 & 33.62 & 36.89 & 35.62 & 29.99 & 33.89 & 29.52 \\
    \hline
    VM-384 (300) + Ours $^*$ & & & & & & & & & & \\
    $\lambda_m$=1e-10 & 0.99 & 32.08 & 34.32 & 25.50 & 33.28 & 36.22 & 34.83 & 29.91 & 33.36 & 29.20 \\
    $\lambda_m$=5e-11 & 1.50 & 32.23 & 34.59 & 25.55 & 33.41 & 36.35 & 35.13 & 29.95 & 33.47 & 29.42 \\
    $\lambda_m$=2.5e-11 & 2.42 & 32.38 & 34.84 & 25.58 & 33.59 & 36.66 & 35.46 & 29.94 & 33.56 & 29.43 \\
    % $\lambda_m$=1e-11 & 3.62 & 32.55 & 35.33 & 25.64 & 33.44 & 36.48 & 35.92 & 29.71 & 34.18 & 29.74 \\
    \hline
    
    VM-384 (500) + Ours $^*$ & & & & & & & & & & \\
    $\lambda_m$=1e-10 & 1.23 & 32.36 & 34.85 & 25.75 & 33.49 & 36.05 & 35.11 & 29.86 & 34.16 & 29.65 \\
    $\lambda_m$=5e-11 & 2.03 & 32.66 & 35.35 & 25.75 & 33.71 & 36.55 & 35.69 & 29.91 & 34.46 & 29.89 \\
    $\lambda_m$=2.5e-11 & 3.36 & 32.77 & 35.49 & 25.78 & 33.78 & 36.79 & 35.85 & 29.94 & 34.56 & 30.01 \\
    % $\lambda_m$=1e-11 & 4.01 & 32.84 & 35.66 & 25.78 & 33.85 & 36.72 & 36.09 & 30.00 & 34.56 & 30.05 \\
    \hline
    
    \end{tabular}}
    \caption{Performance on the NeRF synthetic dataset measured in PSNR.
    The performance of the 32-bit and 8-bit models described in the original paper are represented by the symbols $^\circ$ and $^\bullet$, respectively.
    $^*$ denotes the performance of a quantized model with 8-bit precision.
    The number inside the parenthesis denotes the resolution of one axis of grids.}
    \label{tab:nerf-dataset}
\end{table*}

\begin{table*}[t]
    \centering
    \resizebox{\linewidth}{!}{
    \begin{tabular}{c|c||c||c|c|c|c|c|c|c|c}
    \hline
    Methods & size(MB) & Avg & Bike & Lifestyle & Palace & Robot & Spaceship & Steamtrain & Toad & Wineholder \\
    \hline
    \hline
    KiloNeRF $^\circ$ & $\leq$ 100 & 33.37 & 35.49 & 33.15 & 34.42 & 32.93 & 36.48 & 33.36 & 31.41 & 29.72 \\
    \hline
    VM-192 $^*$ & 17.77 & 36.11 & 38.69 & 34.15 & 37.09 & 37.99 & 37.66 & 37.45 & 34.66 & 31.16 \\ 
    VM-48 $^*$ & 4.53 & 34.95 & 37.55 & 33.34 & 35.84 & 36.60 & 36.38 & 36.68 & 32.97 & 30.26 \\ 
    CP-384 $^*$ & 0.72 & 33.92 & 36.29 & 32.29 & 35.73 & 35.63 & 34.58 & 35.82 & 31.24 & 29.75 \\
    \hline
    VM-192 (300) + Ours $^*$ & & & & & & & & & & \\
    $\lambda_m$=1e-10 & 0.87 & 34.67 & 37.06 & 33.44 & 35.18 & 35.74 & 37.01 & 36.65 & 32.23 & 30.08 \\
    $\lambda_m$=5e-11 & 1.25 & 34.95 & 37.33 & 33.69 & 35.65 & 36.01 & 37.23 & 36.95 & 32.58 & 30.14 \\
    $\lambda_m$=2.5e-11 & 1.88 & 35.11 & 37.49 & 33.75 & 35.94 & 36.23 & 37.45 & 36.92 & 32.87 & 30.23 \\
    % $\lambda_m$=1e-11 & 2.94 & 35.22 & 37.53 & 33.82 & 36.15 & 36.32 & 37.45 & 37.10 & 33.10 & 30.31 \\
    \hline
    VM-192 (500) + Ours $^*$ & & & & & & & & & & \\
    $\lambda_m$=1e-10 & 1.06 & 35.02 & 37.09 & 33.57 & 35.85 & 36.53 & 37.18 & 36.75 & 32.71 & 30.45 \\
    $\lambda_m$=5e-11 & 1.66 & 35.41 & 37.53 & 33.77 & 36.43 & 36.99 & 37.37 & 37.14 & 33.35 & 30.71 \\
    $\lambda_m$=2.5e-11 & 2.63 & 35.63 & 37.70 & 33.96 & 36.86 & 37.15 & 37.60 & 37.26 & 33.77 & 30.71 \\
    % $\lambda_m$=1e-11 & 4.33 & 35.82 & 37.87 & 33.99 & 37.15 & 37.34 & 37.70 & 37.50 & 34.17 & 30.89  \\
    \hline
    VM-384 (300) + Ours $^*$ & & & & & & & & & & \\
    $\lambda_m$=1e-10 & 1.04 & 35.04 & 37.72 & 33.68 & 35.55 & 36.18 & 37.52 & 36.85 & 32.48 & 30.39 \\
    $\lambda_m$=5e-11 & 1.61 & 35.33 & 38.04 & 33.89 & 36.03 & 36.48 & 37.81 & 37.10 & 32.88 & 30.43 \\
    $\lambda_m$=2.5e-11 & 2.69 & 35.57 & 38.27 & 34.09 & 36.37 & 36.81 & 37.93 & 37.24 & 33.22 & 30.59 \\
    % $\lambda_m$=1e-11 & 4.62 & 35.72 & 38.39 & 34.16 & 36.63 & 36.92 & 38.16 & 37.38 & 33.52 & 30.65 \\
    \hline
    VM-384 (500) + Ours $^*$ & & & & & & & & & & \\
    $\lambda_m$=1e-10 & 1.27 & 35.45 & 37.89 & 33.80 & 36.28 & 36.89 & 37.70 & 37.13 & 33.03 & 30.91 \\
    $\lambda_m$=5e-11 & 2.17 & 35.84 & 38.20 & 34.05 & 36.92 & 37.35 & 37.91 & 37.45 & 33.58 & 31.23 \\
    $\lambda_m$=2.5e-11 & 3.77 & 36.13 & 38.53 & 34.26 & 37.32 & 37.71 & 38.15 & 37.73 & 34.08 & 31.27 \\
    % $\lambda_m$=1e-11 & 6.53 & 36.31 & 38.60 & 34.37 & 37.53 & 38.07 & 38.22 & 37.83 & 34.45 & 31.46\\
    \hline
    \end{tabular}}
    \caption{Performance on the NSVF synthetic dataset measured in PSNR. The performance of the 32-bit models described in the original paper are represented by the symbols $^\circ$. $^*$ denotes the performance of a quantized model with 8-bit precision. The number inside the parenthesis denotes the resolution of one axis of grids.}
    \label{tab:nsvf-dataset}
\end{table*}

\begin{table*}[t]
    \centering
    \begin{tabular}{c|c||c||c|c|c|c|c}
    \hline
    Methods & size(MB) & average & Barn & Caterpillar & Family & Ignatius & Truck \\
    \hline
    \hline
    KiloNeRF $^\circ$ & $\leq$ 100 & 28.41 & 27.81 & 25.61 & 33.65 & 27.92 & 27.04 \\
    CCNeRF (CP) $^\circ$ & 4.4 & 27.01 & - & - & - & - & - \\
    \hline
    VM-192 $^*$ & 17.82 & 28.55 & 27.25 & 26.18 & 33.86 & 28.37 & 27.11 \\
    VM-48 $^*$ & 4.52 & 28.06 & 26.77 & 25.46 & 33.06 & 28.24 & 26.77 \\
    CP-384 $^*$ & 0.72 & 27.56 & 26.73 & 24.69 & 32.31 & 27.83 & 26.23  \\
    \hline
    VM-192 (300) + Ours $^*$ &  &  &  &  &  &  & \\
    $\lambda_m$=1e-10 & 0.92 & 27.77 & 26.49 & 25.50 & 32.57 & 28.06 & 26.21 \\
    $\lambda_m$=5e-11 & 1.27 & 27.83 & 26.71 & 25.34 & 32.74 & 28.11 & 26.27 \\
    $\lambda_m$=2.5e-11 & 1.91 & 27.92 & 26.72 & 25.39 & 32.92 & 28.22 & 26.34 \\
    % $\lambda_m$=1e-11 & 3.56 & 27.93 & 26.75 & 25.27 & 33.02 & 28.26 & 26.36 \\
    \hline
    VM-192 (500) + Ours $^*$ &  &  &  &  &  &  & \\
    $\lambda_m$=1e-10 & 1.14 & 27.92 & 26.89 & 25.52 & 32.79 & 28.18 & 26.22 \\
    $\lambda_m$=5e-11 & 1.76 & 28.01 & 26.97 & 25.40 & 33.03 & 28.23 & 26.42 \\
    $\lambda_m$=2.5e-11 & 2.77 & 28.04 & 27.05 & 25.34 & 33.18 & 28.21 & 26.43 \\
    % $\lambda_m$=1e-11 & 4.51 & 28.17 & 27.10 & 25.83 &  33.18 & 28.37 & 26.37 \\
    \hline
    VM-384 (300) + Ours $^*$ &  &  &  &  &  &  & \\
    $\lambda_m$=1e-10 & 1.13 & 28.01 & 26.94 & 25.75 & 32.72 & 28.22 & 26.43 \\
    $\lambda_m$=5e-11 & 1.69 & 28.12 & 27.02 & 25.81 & 32.92 & 28.31 & 26.54 \\
    $\lambda_m$=2.5e-11 & 2.75 & 28.12 & 27.00 & 25.80 & 33.09 & 28.23 & 26.47 \\
    % $\lambda_m$=1e-11 & 4.36 & 28.18 & 27.09 & 25.76 & 33.32 & 28.21 & 26.51 \\
    \hline
    VM-384 (500) + Ours $^*$ &  &  &  &  &  &  & \\
    $\lambda_m$=1e-10 & 1.42 & 28.14 & 27.41 & 25.78 & 32.91 & 28.11 & 26.48 \\
    $\lambda_m$=5e-11 & 2.43 & 28.27 & 27.47 & 25.89 & 33.14 & 28.36 & 26.49 \\
    $\lambda_m$=2.5e-11 & 4.15 & 28.30 & 27.39 & 25.79 & 33.33 & 28.32 & 26.69 \\
    % $\lambda_m$=1e-11 & 6.93 & 28.40 & 27.55 & 26.07 &  33.41 & 28.36 & 26.61 \\
    \hline
    \end{tabular}
    \caption{Performance on the Tanks\&Temples synthetic dataset measured in PSNR. The performance of the 32-bit models described in the original paper are represented by the symbols $^\circ$. $^*$ denotes the performance of a quantized model with 8-bit precision. The number inside the parenthesis denotes the resolution of one axis of grids.}
    \label{tab:Tanks&Temples-dataset}
\end{table*}

\begin{table*}[t]
    \centering
    \resizebox{\linewidth}{!}{
    \begin{tabular}{c|c||c||c|c|c|c|c|c|c|c}
    \hline
    Methods & size(MB) & Avg & Fern & Flower & Fortress & Horns & Leaves & Orchids & Room & T-Rex \\
    \hline
    \hline
    % NeRF $^*$ & ??? & ??? & ??? & ??? & ?? & ? & ? & ? & ? & ? \\
    cNeRF $^\bullet$ & 0.96 & 26.15 & 25.17 & 27.21 & 31.15 & 27.28 & 20.95 & 20.09 & 30.65 & 26.72 \\
    \hline
    PREF $^*$ & 9.34 & 24.50 & 23.32 & 26.37 & 29.71 & 25.24 & 20.21 & 19.02 & 28.45 & 23.67 \\
    \hline
    VM-96 $^*$ & 44.72 & 26.66 & 25.22 & 28.55 & 31.23 & 28.10 & 21.28 & 19.87 & 32.17 & 26.89  \\
    VM-48 $^*$ & 22.40 & 26.46 & 25.27 & 28.19 & 31.06 & 27.59 & 21.33 & 20.03 & 31.70 & 26.54  \\
    CP-384 $^*$ & 0.64 & 25.51 & 24.30 & 26.88 & 30.17 & 26.46 & 20.38 & 19.95 & 30.61 & 25.35  \\
    \hline
    VM-96 (640) + Ours $^*$ & & & & & & & & & \\
    $\lambda_m$=1e-10 & 1.34 & 25.88 & 24.98 & 27.19 & 30.28 & 26.96 & 21.21 & 19.93 & 30.03 & 26.45 \\
    $\lambda_m$=5e-11 & 2.10 & 26.15 & 24.99 & 27.77 & 30.60 & 27.25 & 21.18 & 19.90 & 30.65 & 26.84 \\
    $\lambda_m$=2.5e-11 & 3.20 & 26.25 & 25.05 & 27.94 & 30.75 & 27.48 & 21.08 & 19.76 & 31.19 & 26.77 \\
    % $\lambda_m$=1e-11 & 4.85 & 26.40 & 25.07 & 28.01 & 30.99 & 27.71 & 21.02 & 19.80 & 31.53 & 27.07  \\
    \hline
    VM-96 (1000) + Ours $^*$ & & & & & & & & & \\
    $\lambda_m$=1e-10 & 1.75 & 25.82  & 24.97 & 27.44 & 30.29 & 26.92 & 21.11 & 20.09 & 29.27 & 26.47  \\
    $\lambda_m$=5e-11 & 3.01 & 26.17 & 25.05 & 27.70 & 30.71 & 27.29 & 21.09 & 20.01 & 30.91 & 26.62 \\
    $\lambda_m$=2.5e-11 & 4.76 & 26.30 & 25.08 & 27.76 & 30.89 & 27.49 & 21.14 & 19.99 & 31.23 & 26.85 \\
    % $\lambda_m$=1e-11 & 6.46 & 26.39 & 25.09 & 27.94 & 31.04 & 27.63 & 21.16 & 19.97 & 31.42 & 26.87  \\
    \hline
    VM-192 (640) + Ours $^*$ & & & & & & & & & \\
    $\lambda_m$=1e-10 & 1.73 & 25.98 & 25.18 & 27.47 & 29.66 & 27.47 & 21.11 & 19.71 & 30.47 & 26.79 \\
    $\lambda_m$=5e-11 & 3.01 & 26.46 & 25.12 & 28.16 & 30.81 & 27.88 & 21.07 & 19.77 & 31.61 & 27.28 \\
    $\lambda_m$=2.5e-11 & 5.04 & 26.53 & 25.05 & 28.15 & 30.99 & 28.09 & 20.97 & 19.75 & 31.85 & 27.40 \\
    % $\lambda_m$=1e-11 & 7.49 & 26.64 & 25.02 & 28.19 & 31.20 & 28.27 & 20.90 & 19.80 & 32.23 & 27.52 \\
    \hline
    VM-192 (1000) + Ours $^*$ & & & & & & & & & \\
    $\lambda_m$=1e-10 & 2.43 & 26.15 & 25.18 & 27.74 & 30.22 & 27.47 & 21.24 & 19.98 & 30.56 & 26.79 \\
    $\lambda_m$=5e-11 & 4.57 & 26.43 & 25.22 & 28.06 & 30.78 & 27.71 & 21.23 & 19.92 & 31.61 & 26.93 \\
    $\lambda_m$=2.5e-11 & 7.41 & 26.54 & 25.27 & 28.20 & 31.01 & 27.88 & 21.17 & 20.02 & 31.73 & 27.07 \\
    % $\lambda_m$=1e-11 & 12.94 & 26.49 & 25.17 & 27.96 & 31.03 & 27.79 & 21.18 & 19.92 & 31.89 & 27.01 \\
    \hline
    \end{tabular}}
    \caption{Performance on the LLFF dataset measured in PSNR.
    % The performance of the 32-bit models described in the original paper are represented by the symbols $^\circ$.
    The performance of the 8-bit models described in the original paper are represented by the symbol $^\bullet$.
    $^*$ denotes the performance of a quantized model with 8-bit precision. The number inside the parenthesis denotes the resolution of one axis of grids.}
    \label{tab:llff-dataset}
\end{table*}

% \begin{figure}[t]
% \begin{center}
% \includegraphics[width=0.85\linewidth]{figs/rate_distortion_llff.png}
% \end{center}
% \vspace{-0.5cm}
%   \caption{Rate distortion curves on the LLFF dataset. \textcolor{red}{NERF SHOULD BE ADDED} }
% \label{fig:rate_distortion_llff}
% \vspace{-0.5cm}
% \end{figure}

%%%%%%%%%%% Qualitative results on the real datasets

\end{document}